%% file: arxiv.tex
\documentclass{article}

 \PassOptionsToPackage{numbers, compress}{natbib}

\usepackage[final]{neurips_2021}

\usepackage{caption}
\usepackage{comment}
\usepackage{wrapfig}
\usepackage{amsthm}

\usepackage[toc,page,header]{appendix}
\usepackage{minitoc}

\doparttoc 
\faketableofcontents 

\usepackage[utf8]{inputenc} 
\usepackage[T1]{fontenc}    
\usepackage{hyperref}       
\usepackage{url}            
\usepackage{bbm}
\usepackage{booktabs}       
\usepackage{nicefrac}       

\usepackage{microtype}      
\usepackage[table]{xcolor}
\usepackage{tcolorbox}
\usepackage{hyperref} 
\usepackage{enumitem}
\usepackage{thmtools} 

\usepackage{subfigure}
\hypersetup{
    colorlinks=true,
    citecolor=blue, 
    linkcolor=blue,
    filecolor=magenta,      
    urlcolor=blue,
linktocpage}

\input{math_commands.tex}

\usepackage{mathtools}
\usepackage{algorithm}
\usepackage[nameinlink]{cleveref}

\usepackage{algpseudocode}

\makeatletter
\newenvironment{smalleralign}[1][\small]
 {\par\nopagebreak\leavevmode\vspace*{-\baselineskip}%
  \skip0=\abovedisplayskip
  #1%
  \def\maketag@@@##1{\hbox{\m@th\normalfont\normalsize##1}}%
  \abovedisplayskip=\skip0
  \align}
 {\endalign\ignorespacesafterend}
\makeatother

\title{Neural Auto-Curricula}

\author{Xidong Feng$^{*,1}$, Oliver Slumbers$^{*,1}$,  Ziyu Wan$^2$, Bo Liu$^3$,  \\  \textbf{Stephen McAleer $^4$, Ying Wen$^2$, Jun Wang$^{1}$, Yaodong Yang$^{\dag,5}$} \\ $^1$University College London, $^2$Shanghai Jiao Tong University, \\ $^3$Institute of Automation, CAS,   $^4$University of California, Irvine, \\ $^5$King's College London  }

\begin{document}

\maketitle

\begin{abstract}

When  solving two-player zero-sum games, multi-agent reinforcement learning (MARL) algorithms often create populations of agents where, at each iteration, a new agent is discovered as the best response to a mixture over the opponent population.  Within such a process, the update rules of  \textit{"who to compete with"} (i.e., the opponent mixture) and \textit{"how to beat them"} (i.e., finding best responses)  are underpinned by manually developed game theoretical principles such as fictitious play and Double Oracle.   In this paper\footnote{$^*$Equal contributions.  $^\dag$Corresponding to <yaodong.yang@kcl.ac.uk>.  Code released at  \url{https://github.com/waterhorse1/NAC} }, we introduce a novel framework---Neural Auto-Curricula (NAC)---that leverages meta-gradient descent to automate the discovery of the learning update rule without explicit human design.  Specifically, we parameterise the opponent selection module by neural networks and the best-response module by optimisation subroutines, and update their parameters solely via interaction with the game engine, where both players aim to minimise their exploitability. Surprisingly, even without human design, the discovered MARL algorithms achieve competitive or even better performance with the state-of-the-art population-based game solvers (e.g., PSRO)  on Games of Skill, differentiable Lotto,  non-transitive Mixture Games, Iterated Matching Pennies, and Kuhn Poker. Additionally, we show that NAC is able to generalise from small games to large games, for example training on Kuhn Poker and outperforming PSRO on Leduc Poker. Our work inspires a promising future direction to discover general MARL algorithms solely from data. 
\end{abstract}

\section{Introduction}
\vspace{-0.5em}

Two-player zero-sum games have been a central interest of the recent development of multi-agent reinforcement learning (MARL)  \cite{yang2020overview,baker2019emergent,vinyals2019grandmaster}.  
In solving such games, a MARL agent has a clear objective: to minimise the worst case performance, its exploitability, against any potential opponents. 
When both agents achieve zero exploitability, they reach a Nash equilibrium (NE) \cite{nash1950equilibrium}, a classical solution concept from Game Theory \cite{morgenstern1953theory,deng2021complexity}.  
Even though this objective is straightforward, developing effective algorithms to optimise such an objective often requires tremendous human efforts. One effective approach is through iterative methods, where players iteratively expand a population of agents (a.k.a \emph{auto-curricula} \cite{leibo2019autocurricula}) where at each iteration, a new agent is trained and added to the player's strategy pool.
However, within the auto-curricula process,   it is often non-trivial to design effective  update rules of  \textit{"who to compete with"} (i.e., opponent selections) and \textit{"how to beat them"} (i.e., finding best responses).   
The problem becomes more challenging when one considers additional requirements such as generating agents that have behavioural diversity \cite{yang2021diverse, nieves2021modelling, balduzzi2019open,liu2021unifying} or  generalisation ability \cite{singh2020parrot, oh2020discovering}.
 
 One effective approach to designing auto-curricula is to follow game theoretical principles such as fictitious play \cite{brown1951iterative} and Double Oracle (DO) methods \cite{mcmahan2003planning,dinh2021online}. For example, in the DO dynamics, each player starts from playing a sub-game of the original game where only a restricted set of strategies are available; then, at each iteration, each player will add a best-response strategy, by training against a NE mixture of opponent models at the current sub-game, into its strategy pool.  When an exact best response is hard to compute, approximation methods such as reinforcement learning (RL) methods are often applied \cite{lanctot2017unified}. One of the potential downsides of this approach is, under approximate best response methods, one no longer maintains the theoretical properties of DO and a worthwhile avenue of exploration is to find auto-curricula that are more conducive to these approximation solutions.
  
Apart from following game theoretical principles, an appealing alternative approach one can consider is to automatically discover auto-curricula from data generated by playing games, which can be formulated as a meta-learning problem \cite{finn2017model}. 
However, it remains an open challenge whether it is feasible to discover fundamental concepts  (e.g., NE) entirely based on data. If we can show that it is possible to discover fundamental concepts from scratch, it opens the avenue to trying to discover fundamentally new concepts at a potentially rapid pace. Although encouraging results have been reported in single-agent RL settings showing that it is possible to meta-learn RL update rules, such as temporal difference learning  \cite{oh2020discovering,xu2020meta,xu2018meta}, and the learned update rules can generalise to unseen tasks, we believe discovering auto-curricula in multi-agent cases is particularly hard. 
Two reasons for this are that the discovered auto-curriculum itself directly affects the development of the agent population, and each agent involves an entire training process, which complicates the meta-learning process.  

Albeit challenging, we believe a method capable of discovering auto-curricula without explicit game theoretic knowledge can  potentially open up entirely new approaches to MARL.   
As a result, this paper initiates the study on discovering general-purpose MARL algorithms in two-player zero-sum games. 
Specifically, our goal is to develop an algorithm that learns its own objective (i.e., the auto-curricula) solely from environment interaction, and we offer a meta-learning framework that achieves such a goal. Our solution framework  -- \emph{Neural Auto-Curriculum (NAC)} -- has two promising properties. Firstly, it does not rely on human-designed knowledge about game theoretic principles,  but instead lets the meta-learner decide what the meta-solution concept (i.e., who to compete with) should be during training. This property means that our meta-learner shares the ability of game theoretic principles in being able to accurately evaluate the policies in the population. Secondly, by taking the best-response computation into consideration, NAC can \emph{end to end} offer more \emph{suitable} auto-curricula within the approximate best-response scenario compared with previous approaches; this is particularly important since an exact best-response oracle is not always available in practice.    

Our empirical results show that NAC can discover meaningful solution concepts alike NE, and based on that build effective auto-curricula in training agent populations.   
In multiple different environments, the discovered auto-curriculum  achieves the same performance or better than that of PSRO methods \cite{lanctot2017unified,balduzzi2019open}. 
We additionally evaluate the ability of our discovered meta-solvers to generalise to unseen games of a similar type (e.g., training on Kuhn Poker and testing on Leduc Poker), and show that the auto-curricula found on a simple environment is able to generalise to a more difficult one. To the best of our knowledge, this is the first work that demonstrates the possibility of discovering an entire auto-curriculum in solving two-player zero-sum games, and that the rule discovered from simple domains can be competitive with human-designed algorithms on challenging benchmarks.

\section{Related Work}
\vspace{-0.5em}
Whilst it is theoretically possible to solve for NE in two-player zero-sum games via linear programming (LP) \cite{morgenstern1953theory} in  polynomial time \cite{van2020deterministic}, this is a strictly limited approach. For example, when the action space become prohibitively large, or continuous, using LP becomes intractable; other approximation methods such as fictitious play (FP) \cite{brown1951iterative}, DO  \cite{mcmahan2003planning,dinh2021online} or PSRO \cite{lanctot2017unified, mcaleer2020pipeline, mcaleer2021xdo} methods are required. These methods all follow iterative best-response dynamics where at each iteration, a best response policy is found against a previous aggregated policy (e.g., DO/PSRO applies a sub-game NE, FP applies time-average strategy). Under this general iterative framework, 
  other solution concepts include Rectified NE \cite{balduzzi2019open}, $\alpha$-Rank \cite{muller2019generalized,yang2020alphaalpha}  or no-regret algorithms   \cite{dinh2021online}.  
In this paper, instead of following any existing game theoretic knowledge, we try to discover effective solution concepts solely through environment interactions via meta-learning techniques.  


\begin{figure}[t!]
\vspace{-20pt}
    \centering
    \includegraphics[width=.93\textwidth]{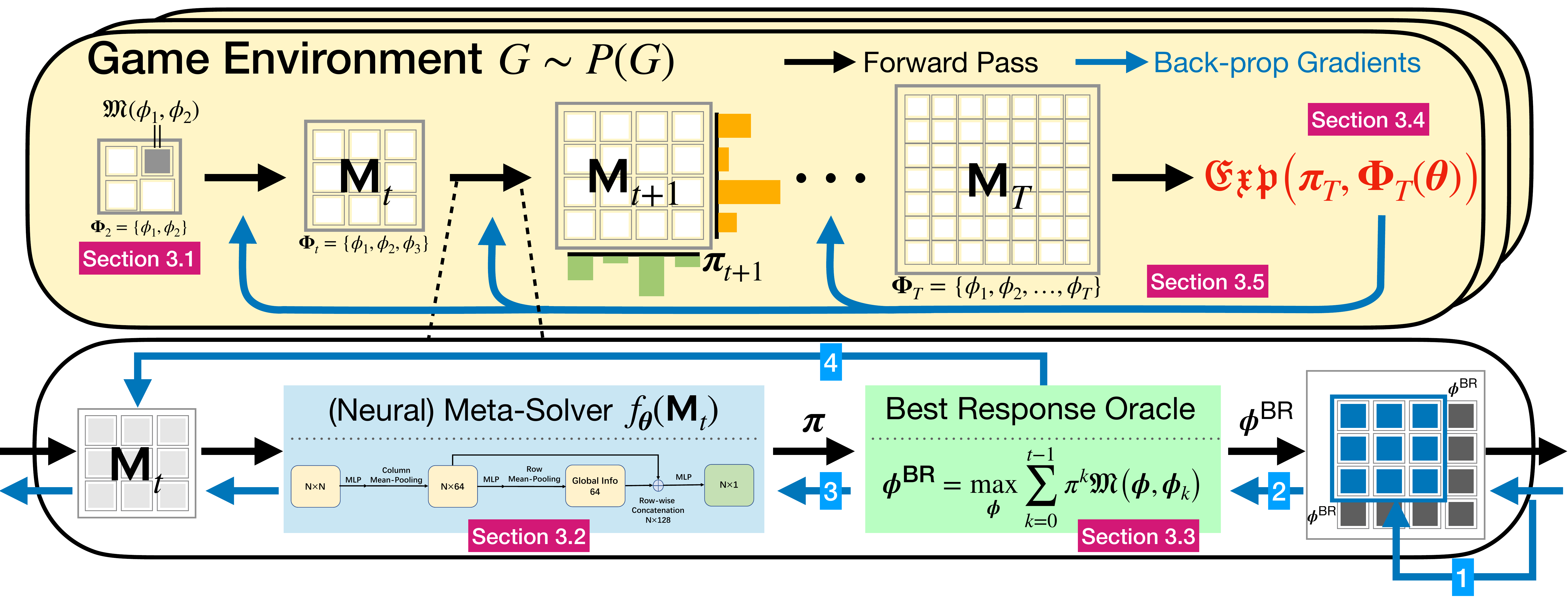}
    \vspace{-5pt}
    \caption{NAC. The top box illustrates the training process where we iteratively expand $\tM_t$ with best responses to $f_\vtheta(\tM_t)$, and then calculate the exploitability (in red). The bottom box illustrates expanding the population by finding $\boldsymbol{\phi}^{\text{BR}}$. The blue arrows show how the gradients pass backwards, from the exploitability through the best-response trajectories. The gradient paths {\color{blue}{[1], [2]}} refer to the two gradient terms in Eq. (\ref{eq:brbp2}) and the  gradient paths {\color{blue}{[3], [4]}} refer to the two gradient paths in Eq. (\ref{eq:brbp}).  }
    \label{fig:main-flow}
    \vspace{-10pt}
\end{figure}

Meta-learning, also known as learning to learn, has recently gained increased attention for successfully improving sample efficiency in regression, classification and RL tasks. 
MAML \cite{finn2017model} meta-learns model parameters by differentiating the learning process for fast adaptation on unseen tasks. 
PROMP \cite{rothfuss2018promp} theoretically analyses the MAML-RL formulation and addresses the biased meta-gradient issue.  
ES-MAML \cite{song2019maml} bypasses the Hessian estimation problem by leveraging evolutionary strategies for gradient-free optimisation. 
RL$^2$ \cite{duan2016rl} and L2RL \cite{wang2016learning}  formulate the meta-learning problem as a second RL procedure but update the parameters at a "slow" pace. 
Recently, there is a trend to meta-learn the algorithmic components of RL algorithms, such as the discount factor \cite{xu2018meta}, intrinsic rewards \cite{zheng2018learning,zheng2020can}, auxiliary tasks \cite{veeriah2019discovery,zhou2020online},  objective functions  for value/policy networks   \cite{bechtle2021meta,xu2020meta,kirsch2019improving}, off-policy update targets \cite{zahavy2020self}, and the bootstrapping mechanism \cite{oh2020discovering}. Their success is built on the \emph{meta-gradient} technique, which leverages gradient descent on the sequence of gradient descent updates resulting from the choice of objective function.  
The meta-learned RL algorithms demonstrate promising generalisation capability in solving different tasks. Apart from meta-gradients, evolutionary strategies (ES) have also been successfully applied \cite{houthooft2018evolved}.  
Our paper, in comparison, has a parallel goal to these prior arts: to discover  MARL algorithms  that are effective for solving various two-player zero-sum games, and to demonstrate their  generalisation ability across different games. 

So far, there have been very few attempts to conduct meta-learning in multi-agent settings \cite{yang2020overview}. MNMPG \cite{shao2021credit} applied meta-gradients for training credit assignment modules for better decomposing the value network \cite{sunehag2017value,yang2020multi}. Meta-PG \cite{al2017continuous} was proposed as a meta-gradient solution to continuous adaptations for non-stationary environments. Building on the opponent modelling technique called LOLA \cite{foerster2017learning,foerster2018dice}, Meta-MAPG \cite{kim2020policy}  extends Meta-PG \cite{al2017continuous} to multi-agent settings by considering both the non-stationarity from the environment and from the opponent's learning process.
Our work rather focuses on automatically discovering  multi-agent auto-curricula through meta-learning, without explicit game theoretic knowledge, that leads to general MARL algorithms for solving two-player zero-sum games. 
\section{Multi-Agent Meta-Learning Framework}
\vspace{-0.5em}
In this section, we discuss our meta-learning framework NAC for discovering multi-agent auto-curricula in two-player zero-sum games. We follow the road-map set out by Fig. (\ref{fig:main-flow}) which illustrates the flow of NAC. The relevant sections for each part of Fig. (\ref{fig:main-flow}) are marked in pink.
\vspace{-0.5em}
\subsection{The Meta-game}
\vspace{-0.5em}
Consider two-player zero-sum games $\mG$\footnote{$\mG$ encapsulates all of the information for a $Player$ to take part in a game (e.g., actions, reward functions).} drawn from some distribution $P(\mG)$, where players have the state-action space $\mS \times \mA$. We start by introducing the notion of an \textit{agent} who is characterised by a policy $\boldsymbol{\phi}$, where a policy is a mapping $\boldsymbol{\phi} : \mS \times \mA \rightarrow [0,1]$ which can be described in both a tabular form or as a neural network. The payoff between two \textit{agents} is defined to be $\mathfrak{M}(\boldsymbol{\phi}_i, \boldsymbol{\phi}_j)$ (i.e., the game engine), and represents the utility to agent $i$, or alternatively, the negative utility to agent $j$. 



Our population-based framework revolves around iterative updates on the meta-game $\tM$. 
At every iteration $t \in T$, a \textit{Player} is defined by a population of fixed \textit{agents} $\boldsymbol{\Phi}_{t} = \boldsymbol{\Phi}_{0} \cup \left\{\boldsymbol{\phi}_1^{\text{BR}}, \boldsymbol{\phi}_2^{\text{BR}}, ... , \boldsymbol{\phi}_{t}^{\text{BR}}\right \}$, where $\boldsymbol{\Phi}_{0}$ is the initial random agent pool and the $\boldsymbol{\phi}^{\text{BR}}_t$ are discussed further in Sec. (\ref{sec:br}). From here, for the sake of notation convenience, we will only consider the \emph{single-population} case where \textit{Players} share the same $\boldsymbol{\Phi}_t$. As such, 
the single population will generate a \textit{meta-game} $\tM_t$, a payoff matrix between all of the \textit{agents} in the population, with individual entries being $\mathfrak{M}(\boldsymbol{\phi}_i, \boldsymbol{\phi}_j) \; \forall \boldsymbol{\phi}_i, \boldsymbol{\phi}_j \in \boldsymbol{\Phi}_{t}.$  

\vspace{-.5em}
\subsection{The Meta-Solver}
\label{meta_solver}
\vspace{-0.5em}
Based on $\tM_{t}$, the \textit{Player} will solve for the meta-distribution $\boldsymbol{\vpi}_{t}$ which is defined as an aggregated \textit{agent} over the fixed agents in $\boldsymbol{\Phi}_{t}$. These meta-solvers in recent work have stuck to human-designed solution concepts such as a uniform distribution \cite{lanctot2017unified} or NE and its variants   \cite{balduzzi2019open,nieves2021modelling,liu2021unifying}, whereas we introduce a method to actively learn these distributions through solely interacting with the game. 
Specifically, we  parameterise our meta-solver via a neural network. This network $f_\vtheta$ with parameters $\vtheta$ is a mapping $f_\vtheta : \tM_t \rightarrow [0,1]^t$ which takes as input a meta-game $\tM_t$ and outputs a \textit{meta-distribution} $\vpi_t = f_\vtheta(\tM_t)$. The output $\vpi_t$ is a probability assignment to each \textit{agent} in the population $\boldsymbol{\Phi}_t$ and, as we are in the \textit{single}-population setting, we do not distinguish between different populations.

There are two characteristics that our network $f_\vtheta$ should have: \textbf{firstly}, it should be able to handle  \emph{variable-length}  inputs as the size of $\tM_t$ is $(t \times t)$, and $f_\vtheta$ outputs a vector of size $(t \times 1)$, where the value of $t$ increments at every iteration. \textbf{Secondly}, $f_\vtheta$ should have both \emph{column-permutation invariance} and \emph{row-permutation equivariance}. Given an input $\tM_t$, $f_\vtheta$ will output the distribution $\vpi_t = f_\vtheta(\tM_t)$ for the row \textit{Player} and it should be equivariant to any permutation of the row index, and be invariant to any permutation of the column index as neither of these actions should affect $\vpi_t$.

Much work has been done on how to maintain permutation invariance/equivariance for neural networks such as Deepsets \cite{zaheer2017deep} for handling tasks defined on sets, or PointNet \cite{qi2017pointnet} for handling 3D point cloud data processing. Our first Multi-Layer Perceptron (MLP) based network is inspired by PointNet, which consists of three components: (1) An MLP over each element $\tM_t^{ij}$ for a non-linear representation, (2) Column Mean-Pooling for column-permutation invariant information aggregation, which will be concatenated to each row and (3) Row-wise MLP for the row-permutation equivariant meta distribution. 
We also offer two alternatives, inspired by a fully Conv1d-based model \cite{long2015fully} and a GRU-based model \cite{chung2014empirical}. 
We refer to Appendix \ref{meta-solver} for detailed architectures of our network $f_\vtheta(\cdot)$. 


\vspace{-.5em}
\subsection{Best Response Oracle}\label{sec:br}
\vspace{-0.5em}
Once a meta-distribution $\vpi_t \in \Delta_{|\boldsymbol{\Phi_{t-1}}|}$ is obtained, the goal is to solve for a best-response against $\vpi_t$ to strengthen the population. Formally,  we define
\vspace{-.5em}
\begin{smalleralign}
\vspace{-5pt}
    \mathfrak{M}(\boldsymbol{\phi}, \langle \vpi_t, \vPhi_{t}   \rangle ) \vcentcolon= \sum_{k=1}^{t}\vpi_{t}^{k}\mathfrak{M}(\boldsymbol{\phi},\boldsymbol{\phi}_k),
\end{smalleralign}
to represent the payoff for agent $\boldsymbol{\phi}$ against the aggregated agent (aggregated by $\vpi_t$) of population $\boldsymbol{\Phi}_{t}$. 
Consequently, we have that the best-response to an aggregated strategy is:
\vspace{-.5em}
\begin{smalleralign}
\vspace{-5pt}
    \boldsymbol{\phi}_{t+1}^{\text{BR}} = \argmax_{\boldsymbol{\phi}} \sum_{k=1}^{t}\vpi_t^k \mathfrak{M}(\boldsymbol{\phi}, \boldsymbol{\phi}_k),
\end{smalleralign}
and the best-response is appended to the population to form a new fixed population, aiming to strengthen the population so as to become less exploitable.  
Depending on the sophistication of games, one may choose appropriate oracles. We consider different oracle implementations in Sec.  
(\ref{sec:meta_solver_upd}). 

\vspace{-.5em}
\subsection{The Learning Objective of Players}
\vspace{-0.5em}
The  goal of NAC is to find an auto-curricula that after $T$ best-response iterations returns a meta-strategy and population, $\langle\vpi_T, \vPhi_{T}\rangle$, that helps minimise the exploitability, written as:
\begin{smalleralign}
\vspace{-5pt}
\label{exp}
 &\min_\vtheta\mathfrak{Exp}(\vpi_T(\vtheta),\vPhi_T(\vtheta)),\text{ where }\mathfrak{Exp}\vcentcolon=  \max_{\boldsymbol{\phi}} \mathfrak{M}\left(\boldsymbol{\phi}, \langle \vpi_{T}, \vPhi_{T} \rangle \right), \\
 &  \ \  \vpi_{T} =  f_{\vtheta}(\tM_T), \vPhi_{T}  = \left\{\boldsymbol{\phi_T^{\text{BR}}}(\vtheta), \boldsymbol{\phi_{T-1}^{\text{BR}}}(\vtheta), ..., \boldsymbol{\phi_{1}^{\text{BR}}}(\vtheta)\right\}. 
    \label{exp2}
\end{smalleralign}
$  \mathfrak{Exp}(\cdot)$ \footnote{In the \textit{multi-}population case, $\mathfrak{Exp}$ expresses the notion of each \textit{Player} having a different population and final meta-strategy, in the single-population case we only need to evaluate the deviation incentive for one population.} represents \emph{exploitability} \cite{davis2014using}, a measure of the incentive to deviate from the meta-strategy over $\langle\vpi_T, \vPhi_{T}\rangle$. 
When exploitability reaches zero, it means one can no longer improve performance. $\boldsymbol{\phi_{t}^{\text{BR}}}(\vtheta)$ in Eq. (\ref{exp2}) shows that each best-response has a dependency on $\vtheta$ since $\boldsymbol{\phi_{t}^{\text{BR}}}$ is influenced explicitly by the curriculum at iteration $t$ and implicitly by all previous curricula. We believe such an objective maximises the generality of our framework on solving different types of zero-sum games. 

NAC can use any oracle, however different considerations must be taken dependent on the choice. In the following sections we will discuss the technicality of directly solving for the meta-gradients of $\vtheta$ with respect to a gradient-descent (GD) oracle and an RL-based oracle. Additionally, we will provide an oracle-agnostic method based on zero-order gradients that allows us to ignore oracle trajectories via Evolutionary Strategies \cite{salimans2017evolution}. Pseudo-code\footnote{Pseudo-code including details of the best-response oracles is shown in Appendix \ref{pseudo}} for these methods is shown in Alg. (\ref{alg:auto-psro-es}).

\vspace{-12pt}

\begin{minipage}{0.98\textwidth}
\vspace{-0pt}
\begin{algorithm}[H]
\caption{Neural Auto-Curricula (NAC) - statements in \textcolor{teal}{teal} refer only to ES-NAC, statements in \textcolor{red}{red} only to standard NAC.}
\label{alg:auto-psro-es}
\begin{algorithmic}[1]
\Require Game distribution $p(\mG)$, learning rate $\alpha$, time window $T$, \textcolor{teal}{perturbations $n$, precision $\sigma$}.
\State Randomly initialise policy pool $\boldsymbol{\phi_{0}}$, Initialise parameters $\vtheta$ of the meta solver $f_\vtheta$.
\For{each training iteration}
\State{\textcolor{red}{Store current model $f_\vtheta$}.} 
\State{\textcolor{teal}{Sample $\boldsymbol{\epsilon}_1,...,\boldsymbol{\epsilon}_n \sim \mathcal{N}(0, I)$ and store $n$ models $f_{(\vtheta + \boldsymbol{\epsilon}_i)}$.}} \Comment{\textit{If using ES, include this step}}
\For{each stored model $f$}
\State Sample games $\{G_{k}\}_{k=1,...,K}$ from $p(\mG)$. 
\For{each game $G_k$}
\For{each  iteration $t$}
\State Compute the meta-policy   $\vpi_{t-1}=f(\tM_{t-1})$. 
\State Compute the best response $\boldsymbol{\phi}_{t}^{\text{BR}}$ by Eq. (\ref{eq:one-step}) or Eq. (\ref{eq:dice}). 
\State Expand the population $\boldsymbol{\Phi_{t}} = \boldsymbol{\Phi_{t-1}} \cup \{\boldsymbol{\phi}_{t}^{\text{BR}}\}$ 
\EndFor
\State Compute $\mathfrak{Exp}_i(\vpi_T, \boldsymbol{\Phi}_T)$ by Eq. (\ref{exp})
\EndFor
\State{Compute the meta-gradient $\textbf{g}_k$ via \textcolor{red}{Eq. (\ref{meta_gradient})} or \textcolor{teal}{Eq. (\ref{es:metagrad})}}
\State Update meta-solver's parameters $\vtheta^{\prime} = \vtheta - \alpha \frac{1}{K} \sum_k{\textbf{g}_k}$.
\EndFor
\EndFor
\end{algorithmic}
\end{algorithm}
\end{minipage}
\vspace{-0.5em}
\subsection{Optimising the Meta-Solver through   Meta-gradients}\label{sec:meta_solver_upd}
\label{meta_objective}
\vspace{-.5em}
Based on the \textit{Player's} learning objectives in Eq. (\ref{exp}), we can optimise the meta-solver as follows:   
\begin{smalleralign}
\vspace{-0pt} 
    \vtheta^* = \argmin_\vtheta \; J \left(\vtheta\right), \; \text{where } J(\vtheta) = \mathbb{E}_{\mG \sim P(\mG)} \big[ \mathfrak{Exp}\left(\vpi, \vPhi |\vtheta, \mG \right)\big]. 
    \label{eq:obj}
    \vspace{-0pt}
\end{smalleralign}
Deriving the (meta-)gradient of $\vtheta$ is non-trivial, which we show in the below Remark  (\ref{remark:gradients}).

\begin{restatable}{remark}{gradients}

\label{remark:gradients}
For a given distribution of game $p(\mG)$, by denoting exploitability at the final iteration  $\mathfrak{M}\left(\boldsymbol{\phi}_{T+1}^{\text{BR}}, \langle \vpi_{T}, \vPhi_{T} \rangle \right)$ as $\mathfrak{M}_{T+1}$, the meta-gradient for $\vtheta$ (see also Fig. \ref{fig:main-flow}) is
\vspace{-.0em}
\begin{smalleralign}
\vspace{-5pt}
\label{meta_gradient}
\nabla_{\vtheta}J(\vtheta) 
    =& \mathbb{E}_\mG\Big[\frac{\partial \mathfrak{M}_{T+1}}{\partial \boldsymbol{\phi}_{T+1}^{\text{BR}}}\frac{\partial \boldsymbol{\phi}_{T+1}^{\text{BR}}}{\partial \vtheta} + \frac{\partial \mathfrak{M}_{T+1}}{\partial \vpi_{T}}\frac{\partial \vpi_{T}}{\partial \vtheta} + \frac{\partial \mathfrak{M}_{T+1}}{\partial \vPhi_{T}}\frac{\partial \vPhi_{T}}{\partial \vtheta}\Big], \text{where}\\
    \label{eq:brbp}
     \frac{\partial \vpi_T}{\partial \vtheta}
= &\frac{\partial f_{\vtheta}(\tM_{T})}{\partial \vtheta} + \frac{\partial f_{\vtheta}(\tM_{T})}{\partial \tM_{T}}\frac{\partial \tM_{T}}{\partial \boldsymbol\vPhi_{T}}\frac{\partial \boldsymbol{\vPhi_{T}}}{\partial \vtheta}, \ 
\frac{\partial \boldsymbol{\phi}_{T+1}^{\text{BR}}}{\partial \vtheta} = \frac{\partial \boldsymbol{\phi}_{T+1}^{\text{BR}}}{\partial \vpi_T}\frac{\partial \vpi_T}{\partial \vtheta} + \frac{\partial \boldsymbol{\phi}_{T+1}^{\text{BR}}}{\partial \vPhi_{T}}\frac{\partial \vPhi_{T}}{\partial \vtheta},\\
\frac{\partial \boldsymbol{\vPhi_{T}}}{\partial \vtheta} =& \left\{\frac{\partial \boldsymbol{\vPhi_{T-1}}}{\partial \vtheta},\frac{\partial\boldsymbol{\phi}_{T}^{\text{BR}}}{\partial \vtheta} \right\},
\label{eq:brbp2}
\end{smalleralign}
\vspace{-.0em}
and Eq. (\ref{eq:brbp2}) can be further decomposed by iteratively applying Eq. (\ref{eq:brbp}) from iteration $T-1$ to $0$.
\vspace{-0pt}
\end{restatable}

The full proof is in Appendix \ref{meta_gradient_theory}. 
 Intuitively, in the forward process,  the population adds $T$ new agents.
 In the backward process, the meta-gradient traverses through the full  $T$ best-response iterations (each iteration may involve many gradient updates) and  back-propagates through all trajectories.  
Therefore, the gradients of ${\partial \boldsymbol{\vPhi_{T}}}/{\partial \vtheta}$
need collecting from $\tM_{T-1}$ to $\tM_1$.  Whilst this is critical in ensuring that every \textit{agent} is influential in optimising $\vtheta$, it introduces computational troubles. Firstly, due to the long-trajectory dependency, computing meta-gradients becomes inefficient due to  multiple Hessian-vector products. 
Secondly, the gradients are susceptible to exploding/vanishing gradients in the same manner as RNNs \cite{hochreiter1997long}. To alleviate these issues, we introduce a truncated version similar to  \cite{oh2020discovering}, where we back-propagate up to a smaller \textbf{window size} (i.e., $n < T$) of  population updates. We shall study the effect of the window size later in Figure \ref{fig:ablation}.  
Notably, the gradients of ${\partial \boldsymbol{\phi}_{t+1}^{\text{BR}}}/{\partial \vPhi_{t}}$ and ${\partial \boldsymbol{\phi}_{t+1}^{\text{BR}}}/{\partial \vpi_{t}}$ in Eq. (\ref{eq:brbp}) also depends on the type of best-response subroutines. In the next section, we demonstrate two types of oracles and show how the meta-gradients are derived accordingly.

\vspace{-.5em}
\subsubsection{Gradient-Descent Best-Response  Oracles} \label{sec:gdbr}
\vspace{-0.5em}
When the payoff function $\mG$ is known and differentiable, one can approximate the best-response through gradient descent (GD).
A one-step GD oracle example is written as:   
\begin{smalleralign}
\vspace{-5pt}
        \boldsymbol{\phi}_{t+1}^{\text{BR}}=\boldsymbol{\phi}_{0}+\alpha \frac{\partial \mathfrak{M}\left(\boldsymbol{\phi}_0, \langle \vpi_t, \vPhi_{t} \rangle \right)}{\partial \boldsymbol{\phi}_0},\label{eq:one-step}
\end{smalleralign}
where $\boldsymbol{\phi}_0$ and $\alpha$ denote the initial parameters and learning rate respectively. 
The backward gradients of one-step GD share similarities with MAML \cite{finn2017model}, which can be written as:
\begin{smalleralign}
\vspace{-5pt}
    \frac{\partial \boldsymbol{\phi}_{t+1}^{\text{BR}}}{\partial \vpi_t} =\alpha\frac{\partial^{2} \mathfrak{M}\left(\boldsymbol{\phi}_{0}, \langle\vpi_{t}, {\vPhi}_{t}\rangle\right)}{\partial \boldsymbol{\phi}_0 \partial \vpi_t}, \ \  \frac{\partial \boldsymbol{\phi}_{t+1}^{\text{BR}}}{\partial \vPhi_{t}} = \alpha \frac{\partial^2 \mathfrak{M}\left(\boldsymbol{\phi}_{0}, \langle\vpi_{t}, \vPhi_{t}\rangle\right)}{\partial \boldsymbol{\phi}_0 \partial \vPhi_{t}}. 
    \label{eq:br_bp}
\end{smalleralign}
We refer to Appendix \ref{gd_meta_gradient} for the specification of Remark (\ref{remark:gradients}) for gradient descent-based oracles.


Though Eq. (\ref{eq:one-step}) and Eq. (\ref{eq:br_bp}) can be easily extended to multi-step GD case, it becomes easily intractable to take the gradient of a computational graph that includes hundreds of gradient updates \cite{song2019maml,rajeswaran2019metalearning,lorraine2020optimizing}. To solve this problem, 
we offer another solution for efficient back-propagation based on the \emph{implicit gradient} method \cite{rajeswaran2019metalearning}, which does not need the full trajectory as a dependency.  
The idea is that,
when we arrive at the best response point such that  ${\partial \mathfrak{M}(\boldsymbol{\phi}_{t+1}^{\text{BR}}, \langle\pi_t, \boldsymbol{\Phi}_{t}\rangle)}/{\partial \boldsymbol{\phi}_t^{\text{BR}}} = 0$, we can apply 
the implicit function theorem and derive the gradient by,
\begin{smalleralign}
\vspace{-5pt}
\frac{\partial\boldsymbol{\phi}_{t+1}^{\text{BR}}}{\partial \vPhi_{t}}
&=  -\left[\frac{\partial^2 \mathfrak{M}(\boldsymbol{\phi}_{t+1}^{\text{BR}}, \langle\vpi_t, \boldsymbol{\Phi_{t}}\rangle)}{\partial \boldsymbol{\phi}_{t+1}^{\text{BR}}\partial {\boldsymbol{\phi}_{t+1}^{\text{BR}}}^ {T}}\right]^{-1}\frac{\partial^{2} \mathfrak{M}(\boldsymbol{\phi}_{t+1}^{\text{BR}}, \langle \vpi_t, \boldsymbol{\Phi_{t}}\rangle)}{\partial \boldsymbol{\phi}_{t+1}^{\text{BR}} \partial \boldsymbol{\Phi_{t}}}.
\label{eq:imp2}
\end{smalleralign}
See  the proof  in Appendix \ref{ig_meta_gradient}. 
Eq. (\ref{eq:imp2}) allows for efficient back-propagation by ignoring the dependency on  trajectories. 
Note that the implicit gradient methods can theoretically  be applied to compute the meta-gradient \emph{w.r.t} RL oracles (next section), but empirically we had little success.   

\vspace{-.5em}
\subsubsection{Reinforcement Learning Best-Response Oracles}
\vspace{-0.5em}
The above GD-based oracles require the pay-off function (i.e., the game engine) to be differentiable. 
Yet, for complex real-world games such as StarCraft \cite{vinyals2019grandmaster}, we have to rely on RL methods to approximate the best-response agent. 
Overall, the RL meta-gradient shares a similar structure to that of the above. The major difference is that we replace the GD terms with Policy Gradient estimator \cite{williams1992simple}. 
Considering the unbiased estimators for the first and the second-order meta-(policy-)gradients, we apply 
Differentiable Monte Carlo Estimator (DICE) \cite{foerster2018dice} in Eq. (\ref{eq:one-step}). DICE is an unbiased higher-order gradient estimator that is fully compatible with automatic differentiation. 
Thanks to DICE, for an RL-based oracle, by regarding the best-response agent $\boldsymbol{\phi}_t^{\text{BR}}$ as  $\boldsymbol{\phi_1}$ and the aggregated agent $\langle\vpi_t, \vPhi_{t}\rangle$ as $\boldsymbol{\phi_2}$ respectively, we obtain the follow equation: 
\begin{smalleralign}
\vspace{-5pt}
     \boldsymbol{\phi}_{1}=\boldsymbol{\phi_}{0}+\alpha \frac{\partial \mathcal{J}^{\mathrm{DICE}}}{\partial \boldsymbol{\phi}_{0}}, \text { where } \mathcal{J}^{\mathrm{DICE}}=\sum_{k=0}^{H-1}\left(\prod_{k^{\prime}=0}^{k} \frac{\vpi_{\boldsymbol{\phi}_{1}}\left(a_{k^{\prime}}^{1} \mid s_{k^{\prime}}^{1}\right) \vpi_{\boldsymbol{\phi}_{2}}\left(a_{k^{\prime}}^{2} \mid s_{k^{\prime}}^{2}\right)}{\perp\left(\vpi_{\boldsymbol{\phi}_{1}}\left(a_{k^{\prime}}^{1} \mid s_{k^{\prime}}^{1}\right) \vpi_{\boldsymbol{\phi}_{2}}\left(a_{k^{\prime}}^{2} \mid s_{k^{\prime}}^{2}\right)\right)}\right) r_{k}^{1},
    \label{eq:dice}
\end{smalleralign}
where $\perp$ refers to the  stop-gradient operator, $r_k^1$ for the reward for agent 1, and $H$ represents the trajectory length.
We refer to Appendix \ref{pg_meta_gradient} for how DICE provides the unbiased first and second-order meta gradient, and the specification of Remark (\ref{remark:gradients}) for RL-based oracles. This RL-based formulation is limited in the fact that it is does not directly extend to SOTA RL techniques such as value-based methods \cite{hessel2018rainbow}, and therefore we next introduce a zero-order method that is able to tackle the case of a non-differentiable pay-off function with any best-response oracle.
\vspace{-2pt}
\subsection{Optimising the Meta-Solver through Evolution Strategies}
\label{sec:es_NAC}
\vspace{-0.5em}
Inspired by the generality of Evolutionary Strategies (ES) \cite{salimans2017evolution}  in optimising black-box functions and ES-based MAML \cite{song2019maml}, we also propose an ES based framework that can cope with any  best-response oracle and underlying game engine. We name our method ES-NAC. 


The ES approach of \cite{salimans2017evolution} states that if we have an objective function $F(\vtheta)$ over a network parameterised by $\vtheta$, we can apply Gaussian perturbations to the parameters  so that  a  gradient estimate of the objective function can be achieved by  $\nabla_\vtheta \mathbb{E}_{\boldsymbol{\epsilon} \sim \mathcal{N}(0,I)}F(\vtheta + \sigma \boldsymbol{\epsilon}) = \frac{1}{\sigma}\mathbb{E}_{\boldsymbol{\epsilon} \sim \mathcal{N}(0,I)}F(\vtheta + \sigma \boldsymbol{\epsilon})\boldsymbol{\epsilon}$. 
In our framework, if we set the objective function $F(\vtheta) = \mathfrak{Exp}_T(\vpi_T, \vPhi_T)$ and $\vpi_T = f_\vtheta(\tM_{T})$,  we can have an objective function as a function of $\vtheta$ which can be perturbed. This allows us to write the gradient of a  surrogate objective  of Eq. (\ref{eq:obj}) as follows:
\vspace{-10pt}
\begin{smalleralign}
    \nabla_\vtheta \hat{J}_\sigma(\vtheta) = \mathbb{E}_{\substack{\mG \sim P(\mG),  \boldsymbol{\epsilon} \sim \mathcal{N}(0,I)}}\left[\frac{1}{\sigma}\big(\mathfrak{Exp}_T(\vpi_T, \vPhi_T)\Big|\vtheta+ \boldsymbol{\epsilon}, \mG \big) \boldsymbol{\epsilon}\right]. 
\label{es:metagrad}
\end{smalleralign}
Additionally, we make use of control variates \cite{liu2019taming} to reduce the variance of the estimator whilst remaining unbiased, for example we apply forward finite differences \cite{choromanski2018structured} whereby the exploitability of the unperturbed meta-solver $f_\vtheta$ is subtracted from the perturbed meta-solver, that is 
\begin{smalleralign}
\vspace{-5pt}
    \nabla_\vtheta \mathbb{E}_{\boldsymbol{\epsilon} \sim \mathcal{N}(0,I)}F\big(\vtheta + \sigma \boldsymbol{\epsilon}\big) = \frac{1}{\sigma}\mathbb{E}_{\boldsymbol{\epsilon} \sim \mathcal{N}(0,I)}\big(F(\vtheta + \sigma \boldsymbol{\epsilon}) - F(\vtheta)\big)\boldsymbol{\epsilon}.
\label{eq:cvar} 
\end{smalleralign}
The key benefit of ES-NAC is that it is agnostic to the best-response oracle choice, as only the final exploitability is required. Unlike the implicit formulation we provided in Sec. (\ref{sec:gdbr}), it is not restricted by the fixed-point condition, which we note is difficult to attain for RL oracles, and therefore may be more widely applicable. This is particularly useful in practice since most games require hundreds of game simulations for each entry in  $\tM$ (e.g., StarCraft \cite{vinyals2019grandmaster}), in which case  we lose the applicability of  either GD or RL-based oracles. We note that Alg. (\ref{alg:auto-psro-es}) encapsulates ES-NAC when the number of perturbations $n > 0$, and that lines in {\color{teal}{teal}} refer only to the ES formulation.

\vspace{-2pt}

\section{Experiments}\label{sec:exps}
\vspace{-0.5em}
We validate the effectiveness of NAC on five  types of zero-sum environments\footnote{To stay self-contained, we provide a detailed description for each game  in Appendix \ref{envir}.} with different levels of complexity.  
They are Games of Skill (GoS) \cite{czarnecki2020real}, differentiable Lotto \cite{balduzzi2019open},  non-transitive mixture game (2D-RPS) \cite{nieves2021modelling}, iterated matching pennies (IMP) \cite{foerster2017learning,kim2020policy} and Kuhn Poker \cite{kuhnpoker}. All selected games are non-trivial to solve as an effective solver has to consider both \emph{transitive} and \emph{non-transitive} dynamics in the policy space \cite{czarnecki2020real,sanjaya2021measuring}.  
The motivation behind our selections is to evaluate the performance of NAC under the different oracles proposed in Sec. (\ref{meta_objective}) and Sec. (\ref{sec:es_NAC}). 
Specifically, we test the gradient descent-oracle (GD) in GoS, Lotto and 2D-RPS and the RL oracle in IMP.
For ES-NAC, we conduct experiments on Kuhn poker \cite{kuhnpoker} with two approximate tabular oracles (V1, V2), an exact tabular oracle\footnote{Training performance for the exact tabular oracle provided in Appendix \ref{ap:kuhn-poker}} and a PPO \cite{schulman2017proximal} oracle. 
We conduct the experiments on multiple random seeds for NAC and the details of how we conduct meta-testing on baseline algorithms and NAC are reported in Appendix \ref{Meta-testing}. More details of all of the applied oracles and their hyper-parameters are in Appendix \ref{hyper}, and details of the baseline implementations are in Appendix \ref{additional_implementaion}.


We select the baselines to be vanilla self-play (i.e., best responding to the latest agent in the population) and  the PSRO variants, including  PSRO \cite{lanctot2017unified}, PSRO-Uniform \cite{balduzzi2019open} (equivalent to Fictitious Play \cite{brown1951iterative}) and PSRO-rN \cite{balduzzi2019open}. 
Their implementations can be found in OpenSpiel \cite{lanctot2019openspiel}.  
We believe these methods offer strong benchmarks for NAC since they are all underpinned by game theoretic principles, and NAC tries to  discover solution algorithms purely from data.  
Results are presented in the form of answering five critical questions \emph{w.r.t} the effectiveness of NAC.

\begin{figure}[t!]
\vspace{-10pt}
    \centering
    \includegraphics[width=\linewidth]{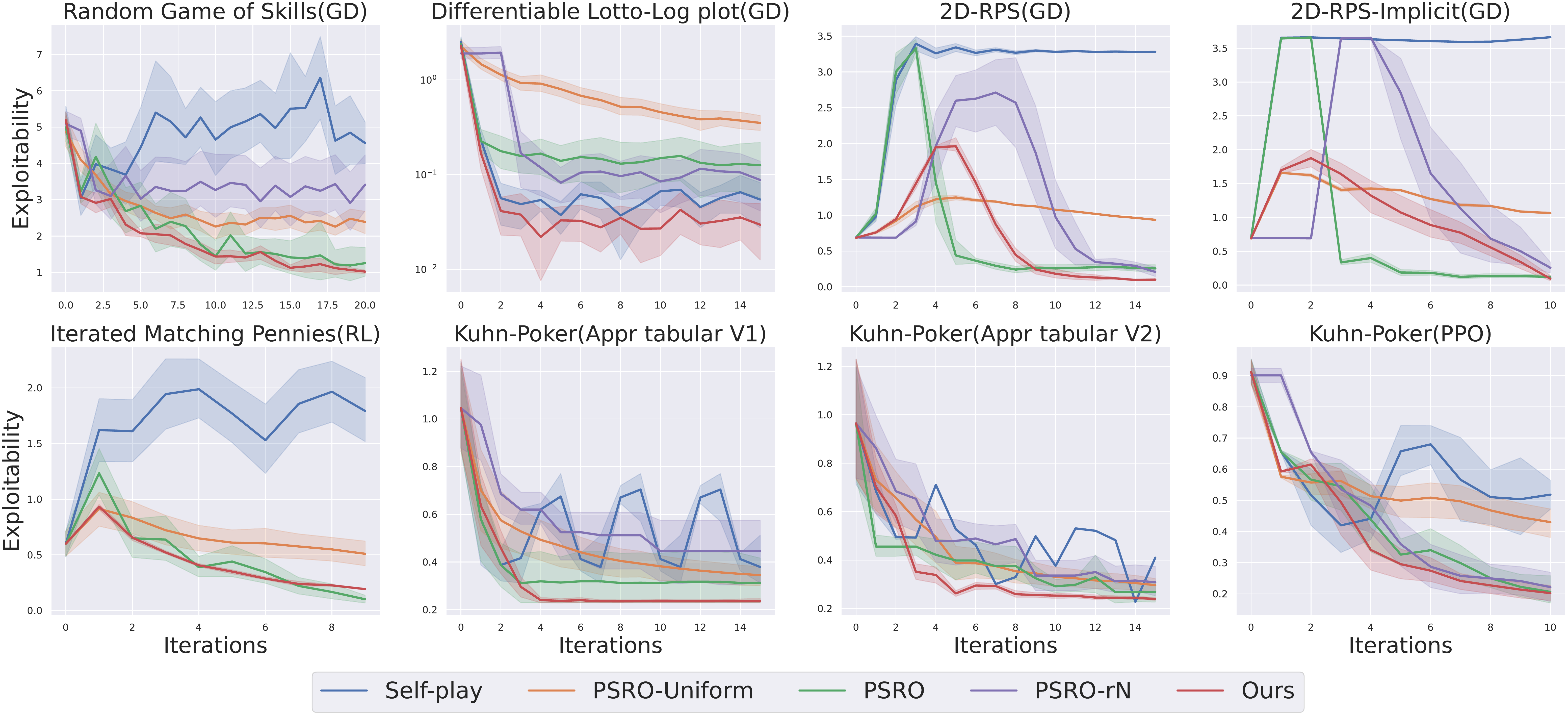}
    \vspace{-.5em}
    \caption{Exploitability results on  five different environments with differing best-response oracles.  NAC performs at least as well as the best baseline in all settings, and often outperforms the PSRO baselines. Settings of adopted oracles in each game can be found in Appendix {\ref{hyper}}. }
    \label{fig:q1exps}
    \vspace{5pt}
\end{figure}

\textbf{Question 1. } \textbf{\emph{How does NAC perform in terms of exploitability on different games?}}

Firstly, we are interested in whether NAC can learn an auto-curricula that can solve games effectively. In order to characterise performance, we measure the  exploitability, Eq.  (\ref{exp}), of NAC and compare it against other baselines.
Surprisingly, results in Fig. (\ref{fig:q1exps}) suggest that, by learning an effective meta-solver, NAC is able to solve games without explicit game theoretic solution concepts. 
In particular, NAC performs at least as well as PSRO (only slightly worse than PSRO in IMP), and in multiple games outperforms PSRO. 
We notice that NAC performs better in the presence of \textit{approximate} best-responses.  
 One explanation is that Double Oracle relies upon a \textit{true} best-response oracle to guarantee convergence, when it comes to PSRO where only \emph{approximate} best responses are available,  the principles of sub-game NE may not necessarily fit with PSRO anymore. 
In contrast, NAC considers the outcomes of approximate best-responses in an end-to-end  fashion; therefore, the auto-curricula for each player tends to be adaptive, leading to the development of a stronger  population.  
Overall, we believe these results suggest a promising avenue of research for solving larger-scale games (e.g., StarCraft \cite{vinyals2019grandmaster} or XLand \cite{team2021open} )   with no  exact best responses available and no prior game theoretic solution concepts  involved.    

\textbf{Question 2. } \textbf{\emph{What does the learned curricula (i.e., $f_{\boldsymbol{\theta}}(\tM_t)$) look like?}}\label{q:vis}
\begin{figure*}[htbp]
\vspace{-0pt}
 		\centering
 		\begin{minipage}[t]{\linewidth}
		\centering
		\includegraphics[width=\linewidth]{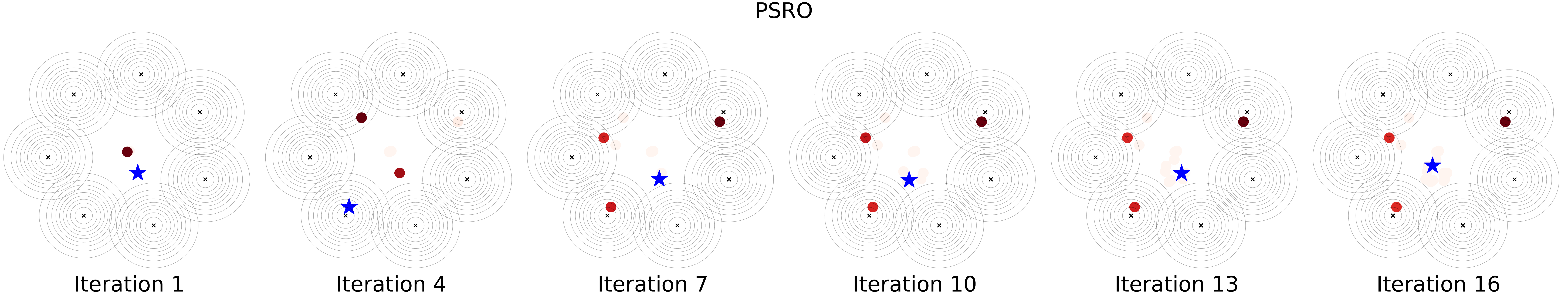}
		\label{fig:hor_2figs_2cap_1}
	\end{minipage}
	\\
	\begin{minipage}[t]{\linewidth}
		\centering
		\includegraphics[width=\linewidth]{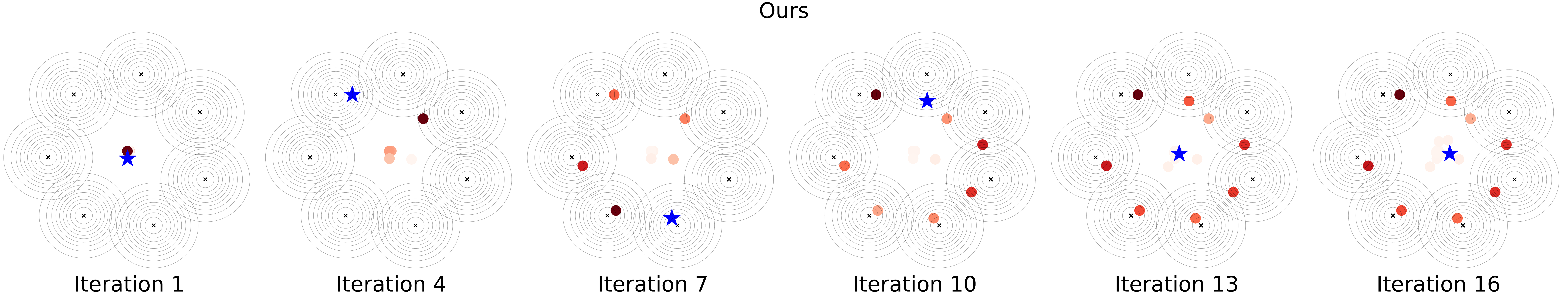}
		\label{fig:hor_2figs_2cap_2}
	\end{minipage}
	\vspace{-.5em}
 		\caption{Visualisations of curricula on 2D-RPS. Red points denote the meta-solver output, and darker refers to higher probability in $\vpi$. The blue star denotes the latest best-response. 
 		To achieve low exploitability, one needs to climb up and explore each Gaussian. 
 		PSRO fails to explore fully, where as NAC creates an effective curricula to explore all modes and assigns each of them high weights. 
 		}
 		\label{visualization}
 		\vspace{0pt}
 	\end{figure*}  

To address this question\footnote{We also visualise Kuhn-Poker policy for understanding how NAC works in Appendix \ref{visualisation}}, we visualise the auto-curricula on the 2D-RPS game between PSRO and NAC in Fig. (\ref{visualization}). 
PSRO is successful in climbing up some Gaussians before iteration $7$;  however, it fails to offer an effective auto-curricula that can lead it to discover other Gaussians. PSRO  fails to select an auto-curricula that takes into consideration whether the best-response oracle is capable of learning a \emph{suitable} best-response. 
This result is inline with \cite{nieves2021modelling} and we believe it is because the approximate best responses may lead to a local optimum in the policy space for PSRO-type methods.   
In contrast,  NAC adaptively generates an auto-curricula which is more \emph{suitable} for approximate best-responses, as evidenced by a wide spread of agents over the plane, and lower exploitability. 

\textbf{Question 3. }  \textbf{\emph{Does back-propogation through multiple best-response iterations  help the training?}}

\begin{figure*}[t]
\vspace{-5pt}
 		\centering
 		\begin{minipage}[t]{0.95\linewidth}
		\centering	\includegraphics[width=\linewidth]{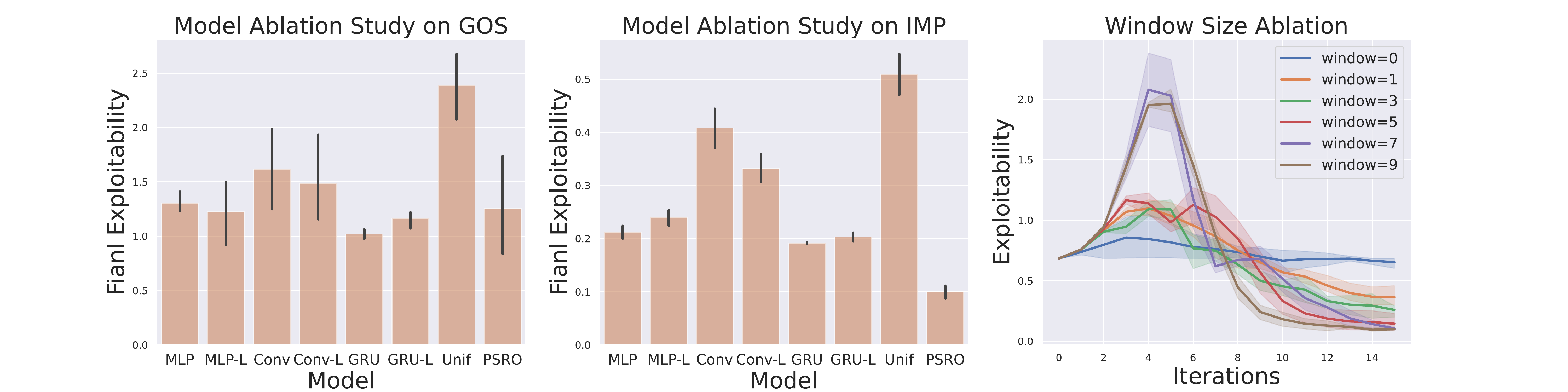}
		\vspace{-1.5em}
		\caption{(a), (b) The effect of different neural architectures on the exploitability in GoS and IMP games. (c) The effect of window size on the exploitability in 2D-RPS game.}
        \label{fig:ablation}
	\end{minipage}
	\vspace{-.0em}
 \end{figure*}

As shown by the blue lines in Fig. (\ref{fig:main-flow}), the backward meta-gradient will propagate  through multiple iterations of best-response processes. To demonstrate its effects, in Fig. (\ref{fig:ablation}c) we conduct an ablation study on NAC by varying the how many best-response iterations (i.e., the window size) we consider, by controlling how many agents are added into the population before computing the meta-gradient. A window size of 0 refers to the setting where we completely detach the gradients of the best-response process. 
We can see that NAC achieves lower exploitability when considering multiple best-response iterations, which reflects the effectiveness of NAC in offering a more \emph{suitable} and appropriate curricula to strengthen the whole population. 

\textbf{Question 4.}  \textbf{\emph{How is NAC affected by the architecture and capacity of meta-solver?}}

In Sec. (\ref{meta_solver}), we provide several neural architectures for the meta-solver. Thus, to understand the effect of these different architectures, we conduct an ablation study on GoS and IMP. We specify six different models by varying both the architecture and the size. 
Results in Fig. (\ref{fig:ablation}a, \ref{fig:ablation}b) show that, firstly, the permutation invaraiance/equivariance is not a necessary property, as the GRU-based models achieve great performance. Secondly, the effect of the meta-solver's architecture is heavily dependent on the game. The performance of all three models are comparable for GOS while only MLP and GRU work well for IMP. Different games may need different meta-solver's architecture and GRU-based meta-solver tend to work better. In addition, the increase of network capacity does not correspond to performance improvement. We refer the reader to Appendix (\ref{ap:hyper}) for details of our model choices for different games.

\vspace{20pt}
\textbf{Question 5. }  \textbf{\emph{What is the generalisation ability of the neural meta-solver by NAC?}}
\begin{figure*}[t]
 		\centering
 		\begin{minipage}[t]{1.0\linewidth}
		\centering
		\includegraphics[width=\linewidth]{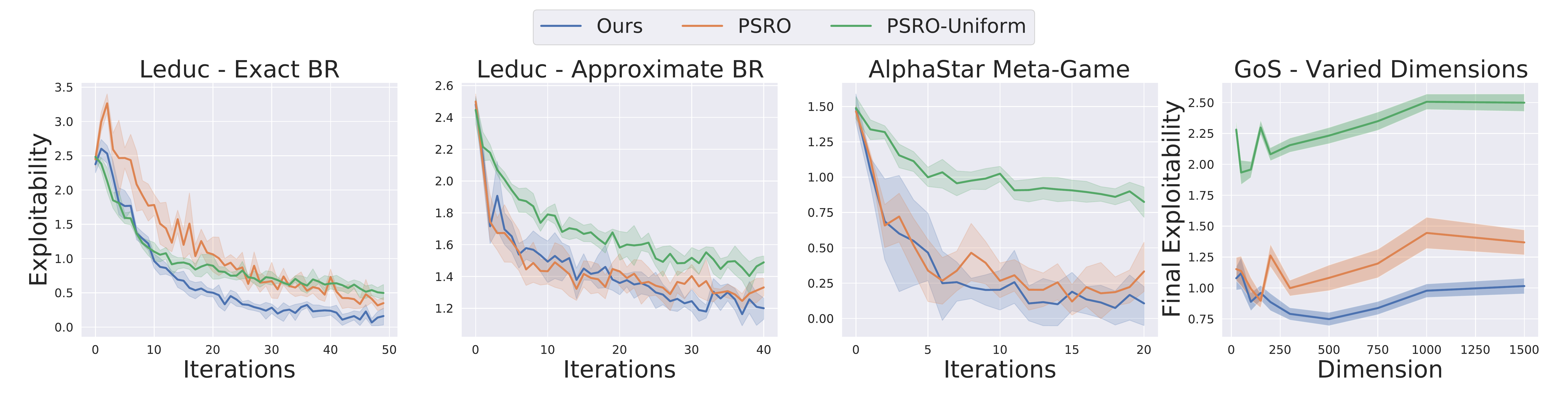}
		\caption{(a) Exploitability when trained on Kuhn Poker with an exact tabular BR oracle using ES-NAC and tested on Leduc Poker. (b) Same as (a) with approximate tabular BR V2 (c) Exploitability when trained on GoS with a GD oracle and tested on the AlphaStar meta-game from \cite{czarnecki2020real} (d) Final exploitability when trained on 200 Dimension GoS and tested on a variety of dimension size GoS.}
        \label{fig:gen}
	\end{minipage}
	\vspace{-.0em}
 \end{figure*}

The  most promising aspect of NAC is that the neural auto-curricula (i.e., meta-solvers) have the ability to generalise to different out-of-distribution games. This is particularly impactful, as it allows for training on simpler games and then deploying them on larger, more difficult games. 
We test the generalisation capability of NAC in two settings. First, we take our meta-solver trained over $200$ dimensional GoS and test on new unseen GoS of varying dimension. We consider this to be the most direct way of ascertaining whether the neural meta-solvers are able to generalise to larger, more difficult games, and whether the in-task performance still holds out-of-task. Fig. (\ref{fig:gen}d) plots the final exploitability after 20 PSRO iterations against the dimension of the GoS, and noticeably, NAC still outperforms the PSRO baselines in all dimensions larger than the training dimension. Additionally, we test our trained meta-solver on the AlphaStar meta-game generated by \cite{czarnecki2020real}\footnote{We provide the results on the other tens of meta-games from \cite{czarnecki2020real} in Appendix \ref{ap:gener}} in Fig. (\ref{fig:gen}c), which is also considered to be a form of a GoS. Interestingly, our meta-solver is able to perform well on a GoS that is outside of the task distribution and therefore has a different type of underlying dynamics.  

Secondly, we introduce an example of our meta-solver showing the ability to scale-up to different games, namely we train on the Kuhn Poker environment ($2^{12}$ pure strategies) and test on the Leduc Poker environment ($3^{936}$ pure strategies). As shown in Fig. (\ref{fig:gen}a, \ref{fig:gen}b) the trained meta-solver is able to outperform the PSRO algorithms when used on Leduc Poker, which suggests NAC enjoys effective generalisation abilities for both an exact best-response oracle and an approximate best-response oracle. We hypothesise that, whilst Leduc Poker is different from Kuhn Poker, the "Poker" nature of both games means they encapsulate similar dynamics, allowing our meta-solver to perform favourably. 
\vspace{-1.5em}
\section{Conclusion}
\vspace{-0.5em}
We introduce a method for discovering  auto-curricula on two-player zero-sum games based on meta-learning. To our best knowledge, we are the first to show that it is entirely possible to perform as well as solutions underpinned in game-theoretic concepts that are designed through human insights, without any active design of the auto-curricula itself. In particular, we show that our NAC method can learn in small games and generalise to larger games, more difficult games that follow a similar underlying structure. We believe this initiates an exciting and promising research area in which large-scale difficult games can be solved effectively by training on simplified versions of the game. 
\bibliographystyle{plain}
\bibliography{references}

\clearpage
\appendix

\addcontentsline{toc}{section}{Appendix} 
\part{{\Large{Supplementary Material for \emph{"Neural  Auto-Curricula in Two-player Zero-sum Games"}}}} 
\parttoc

\newpage
\section{Meta-solver Architecture}
\label{meta-solver}
In this section, we recap the meta-solver properties that we need and illustrate how we designed models to achieve them. There exist two properties the model should have.
\begin{itemize}[leftmargin=*]
\item The model should handle a variable-length matrix input.
\item The model should be subject to row-permutation equivariance and column-permutation invariance.

\end{itemize}
Three different techniques can be utilised to achieve the first property, which also corresponds to the three different models we propose: MLP based, Conv1d based and GRU based model. If not specifically mentioned, we utilise ReLU as the activation function for all MLP used in our meta-solver.

\subsection{MLP-based Meta-Solver}
 \begin{figure}[tbh]
     \centering
     \includegraphics[width=\linewidth]{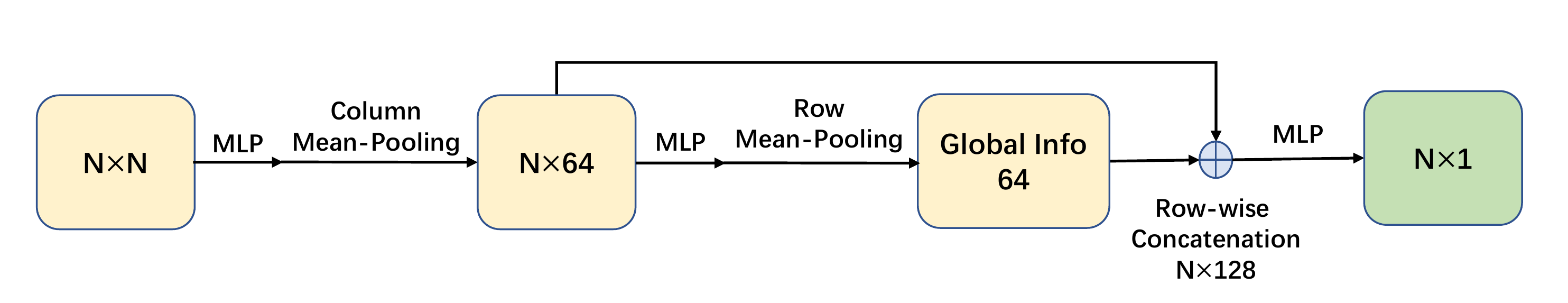}
     \caption{MLP based Meta-Solver}
     \label{fig:mlp}
\end{figure}
The first model is based on MLP. Inspired by PointNet\cite{qi2017pointnet}, we utilise MLP + pooling operation + row-wise operation to handle variable-length matrix inputs and permutation invariance/equivariance. The first MLP + Column Mean-Pooling operation generates row-wise features: $N\times N \rightarrow N\times N \times 64\rightarrow N\times 64$. Then the model transforms it to global matrix information by MLP + Row Mean-Pooling operation: $N \times 64\rightarrow N\times 64\rightarrow 64$. Finally, the model conducts row-wise concatenation between row-wise features and the global matrix information, and transforms it with the last MLP for the final output: $N \times (64+64)\rightarrow N\times 128\rightarrow N\times1$. 

The MLP-based model successfully satisfies the properties we need. However, empirically we find that it does not always perform well within our training framework. We empirically find out even if the model violates the second property, it can still work well. We believe this is because there exists some particular patterns of meta distribution in the PSRO iterations, so even if the model is not a generally proper meta-strategy solver, it can still work well under PSRO. Next, we will detail Conv1d-based and GRU-based models which are not completely subject to the second property.

\subsection{Conv1D-based Meta-Solver}
 \begin{figure}[tbh]
     \centering
     \includegraphics[width=\linewidth]{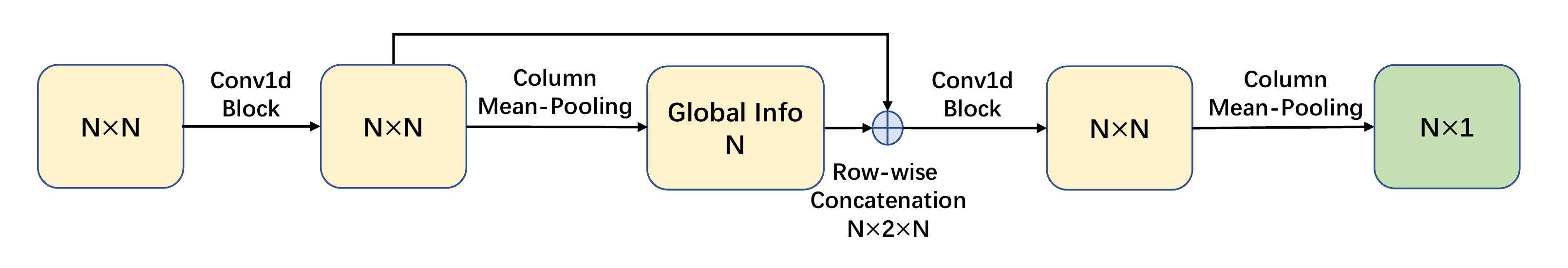}
     \caption{Conv1D based Meta-Solver}
     \label{fig:conv1d}
\end{figure}
Our second model is Conv1d based. To satisfy variable-length matrix inputs, we make use of a Fully Convolutional Neural Network \cite{long2015fully} with Conv1d on the row vectors $\tM_{i}$ which, by construction, with particular kernel and padding size will not decrease the feature size in the forward pass. The procedure is shown as follows. Firstly, the model has multiple Conv1d-LeakyReLU layers (as a Conv1d block) for generating row-wise features: $N\times N\rightarrow N\times N$. Then similar column Mean-Pooling and row-wise concatenation are utilised to achieve global matrix information: $N\times N\rightarrow N; N\times (1+1) \times N \rightarrow N\times 2 \times N$. The final Conv1d block + Column Mean-Pooling operation gives the final prediction result: $N\times 2 \times N \rightarrow N\times N \rightarrow N\times 1$.

Note that Conv1d-based model follows property 1 and row permutation equivariance. However, it violates the column permutation invariance property since we conduct Conv1d operation on column vectors.

\subsection{GRU-based Meta-Solver}
 \begin{figure}[t]
     \centering
     \includegraphics[width=\linewidth]{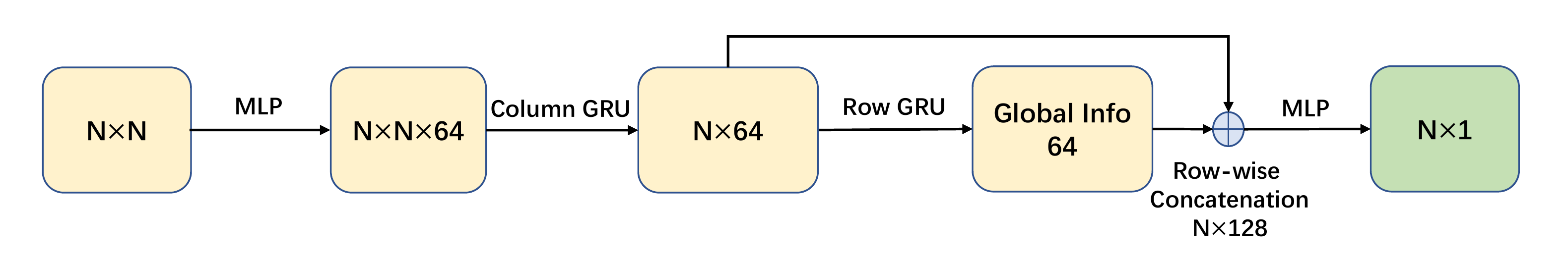}
     \caption{GRU based Meta-solver}
     \label{fig:my_label}
\end{figure}

The final model is based on GRU, which can take in variable-length input sequences. To achieve a variable-length matrix input, we utilise GRU on both the column vectors and row vectors. The procedure is shown as follows. Firstly, the model utilises MLP + column GRU to aggregate the column vector for row-wise features: $N\times N\rightarrow N \times N \times 64 \rightarrow N \times 64$. With similar row GRU + row-wise concatenation + MLP, the model gets the final result: $N\times 64\rightarrow 64; N\times (64+64)\rightarrow N\times 1$

Note that the GRU-based model only follows property 1. With GRU plugged in, it cannot hold both row permutation equivariance and column permutation invariance.

\section{Proof of \Cref{remark:gradients}}
\label{meta_gradient_theory}

\gradients*

\begin{proof}[Proof of \Cref{remark:gradients}]
Here we will only consider the \textit{single-population} case, which is the same as in the main paper for notation convenience. Note that the whole framework can be easily extended to the \textit{multi-population} case. Firstly, we illustrate how the forward process works.

Assume we have $t$ policies in the policy pool $\boldsymbol{\Phi}_{t}$ at the beginning of iteration $t+1$.
\begin{smalleralign}
    \tM_t = \{\mathfrak{M}(\boldsymbol{\phi}_i, \boldsymbol{\phi}_j)\} \; \forall \boldsymbol{\phi}_i, \boldsymbol{\phi}_j \in \boldsymbol{\Phi}_{t}
\end{smalleralign}

We generate the curricula (meta distribution) by meta-solver $f_\vtheta$.

\begin{smalleralign}
    \vpi_t = f_{\vtheta}(\tM_t)
\end{smalleralign}

Then we utilise best-response optimisation w.r.t the mixed meta-policy for the new policy which is added into the policy pool.

\begin{smalleralign}
\mathfrak{M}(\boldsymbol{\phi}, \langle \vpi_t, \vPhi_{t}   \rangle ) \vcentcolon&= \sum_{k=1}^{t}\vpi_{t}^{k}\mathfrak{M}(\boldsymbol{\phi},\boldsymbol{\phi}_k)\\
    \boldsymbol{\phi_{t+1}^{\text{BR}}} &= \mathop{\arg\max}_{\boldsymbol{\phi}} \mathfrak{M}\left(\boldsymbol{\phi}, \langle \vpi_{t}, \vPhi_{t} \rangle \right)\\
    \boldsymbol{\Phi}_{t+1} &= \boldsymbol{\Phi}_{t} \cup \boldsymbol{\phi_{t+1}^{\text{BR}}} 
\end{smalleralign}

After the final iteration $T$, we get the policy pool $\boldsymbol{\Phi}_{T}$ and calculate the exploitability of the final meta-policy:

\begin{smalleralign}
\tM_T &= \{\mathfrak{M}(\boldsymbol{\phi}_i, \boldsymbol{\phi}_j)\} \; \forall \boldsymbol{\phi}_i, \boldsymbol{\phi}_j \in \boldsymbol{\Phi}_{T}\\
\vpi_T &= f_{\vtheta}(\tM_T)\\
    \boldsymbol{\phi_{T+1}^{\text{BR}}} &= \mathop{\arg\max}_{\boldsymbol{\phi}} \mathfrak{M}\left(\boldsymbol{\phi}, \langle \vpi_{T}, \vPhi_{T} \rangle \right)\\
    \mathfrak{Exp}&=\mathfrak{M}\left(\boldsymbol{\phi_{T+1}^{\text{BR}}}, \langle \vpi_{T}, \vPhi_{T}\rangle \right)
\end{smalleralign}

Given a distribution $P(\mG)$ over game $\mG$, the meta-gradient for $\vtheta$ can be derived by applying the chain rule:
\begin{smalleralign}
    \nabla_{\vtheta}J(\vtheta) 
    &= \mathbb{E}_\mG\Big[\frac{\partial \mathfrak{M}_{T+1}}{\partial \boldsymbol{\phi}_{T+1}^{\text{BR}}}\frac{\partial \boldsymbol{\phi}_{T+1}^{\text{BR}}}{\partial \vtheta} + \frac{\partial \mathfrak{M}_{T+1}}{\partial \vpi_{T}}\frac{\partial \vpi_{T}}{\partial \vtheta} + \frac{\partial \mathfrak{M}_{T+1}}{\partial \vPhi_{T}}\frac{\partial \vPhi_{T}}{\partial \vtheta}\Big], \text{where}\\
    \label{eq:pi_theta}
     \frac{\partial \vpi_T}{\partial \vtheta}
&=\frac{\partial f_{\vtheta}(\tM_{T})}{\partial \vtheta} + \frac{\partial f_{\vtheta}(\tM_{T})}{\partial \tM_{T}}\frac{\partial \tM_{T}}{\partial \boldsymbol\vPhi_{T}}\frac{\partial \boldsymbol{\vPhi_{T}}}{\partial \vtheta}, \\
\label{eq:br_theta}
\frac{\partial \boldsymbol{\phi}_{T+1}^{\text{BR}}}{\partial \vtheta} &= \frac{\partial \boldsymbol{\phi}_{T+1}^{\text{BR}}}{\partial \vpi_T}\frac{\partial \vpi_T}{\partial \vtheta} + \frac{\partial \boldsymbol{\phi}_{T+1}^{\text{BR}}}{\partial \vPhi_{T}}\frac{\partial \vPhi_{T}}{\partial \vtheta},\\
\frac{\partial \boldsymbol{\vPhi_{T}}}{\partial \vtheta} &= \left\{\frac{\partial \boldsymbol{\vPhi_{T-1}}}{\partial \vtheta},\frac{\partial\boldsymbol{\phi}_{T}^{\text{BR}}}{\partial \vtheta} \right\}.
\label{eq:phi_theta}
\end{smalleralign}

\end{proof}

Note that Eq. (\ref{eq:phi_theta}) can be further decomposed by iteratively applying Eq. (\ref{eq:pi_theta}) and Eq. (\ref{eq:br_theta}), which means the gradients will backpropagate through multiple iterations. The whole process is similar to the backpropagation through time (BPTT) process in RNN.

In the following section, we detail how the gradient is calculated with two different best-response oracles - a Gradient-Descent oracle and a Reinforcement Learning oracle, in particular showing how we take meta-gradients for both.

\subsection{Gradient-Descent Best-Response with direct Meta-gradient}
\label{gd_meta_gradient}
For a GD based best-response oracle, the payoff function of the game is differentiable, so we can directly obtain gradients $\frac{\partial \mathfrak{M}_{T+1}}{\partial \boldsymbol{\phi}_{T+1}^{\text{BR}}}, \frac{\partial \mathfrak{M}_{T+1}}{\partial \vpi_{T}}, \frac{\partial \mathfrak{M}_{T+1}}{\partial \vPhi_{T}}$ by automatic differentiation.

An example for a GD oracle with one gradient-descent step:   
\begin{smalleralign}
\vspace{-5pt}
\boldsymbol{\phi}_{t+1}^{\text{BR}}=\boldsymbol{\phi}_{0}+\alpha \frac{\partial \mathfrak{M}\left(\boldsymbol{\phi}_0, \langle \vpi_t, \vPhi_{t} \rangle \right)}{\partial \boldsymbol{\phi}_0},\label{ap:eq:one-step}
\end{smalleralign}
where $\boldsymbol{\phi}_0$ and $\alpha$ denote the initial parameters and learning rate respectively. The backward gradients of one-step GD share similarities with MAML \cite{finn2017model}, which can be written as:
\begin{smalleralign}
\vspace{-5pt}
    \frac{\partial \boldsymbol{\phi}_{t+1}^{\text{BR}}}{\partial \vpi_t} =\alpha\frac{\partial^{2} \mathfrak{M}\left(\boldsymbol{\phi}_{0}, \langle\vpi_{t}, {\vPhi}_{t}\rangle\right)}{\partial \boldsymbol{\phi}_0 \partial \vpi_t}, \ \  \frac{\partial \boldsymbol{\phi}_{t+1}^{\text{BR}}}{\partial \vPhi_{t}} = \alpha \frac{\partial^2 \mathfrak{M}\left(\boldsymbol{\phi}_{0}, \langle\vpi_{t}, \vPhi_{t}\rangle\right)}{\partial \boldsymbol{\phi}_0 \partial \vPhi_{t}}. 
    \label{ap:eq:br_bp}
\end{smalleralign}

Eq. (\ref{ap:eq:one-step}) and (\ref{ap:eq:br_bp}) can be easily extended to situations where we take a few gradient steps. Eq. (\ref{eq:phi_theta}) can be calculated iteratively by calling for the previous gradient terms,

\begin{smalleralign}
\frac{\partial \boldsymbol{\vPhi_{T}}}{\partial \vtheta} &= \left\{\frac{\partial\boldsymbol{\phi}_{1}^{\text{BR}}}{\partial \vtheta}, \frac{\partial\boldsymbol{\phi}_{2}^{\text{BR}}}{\partial \vtheta}, ...,\frac{\partial\boldsymbol{\phi}_{T}^{\text{BR}}}{\partial \vtheta} \right\}\nonumber\\
&=\left\{\alpha\frac{\partial^{2} \mathfrak{M}\left(\boldsymbol{\phi}_{0}, \langle\vpi_{t}, {\vPhi}_{t}\rangle\right)}{\partial \boldsymbol{\phi}_0 \partial \vpi_t}\frac{\partial\vpi_t}{\partial\vtheta}+\alpha \frac{\partial^2 \mathfrak{M}\left(\boldsymbol{\phi}_{0}, \langle\vpi_{t}, \vPhi_{t}\rangle\right)}{\partial \boldsymbol{\phi}_0 \partial \vPhi_{t}}\frac{\partial\vPhi_{t}}{\partial\vtheta}\right\}_{t\in\{1,2,...,T\}}
\end{smalleralign}.
\subsection{Gradient-Descent Best-Response with Implicit Gradient based Meta-gradient}
\label{ig_meta_gradient}

The direct meta-gradient formulation above becomes easily intractable when the computational graph including hundreds of gradient updates. Thus, here we offer another alternative based on implicit gradients for efficient meta-gradient backpropagation. The main issue here is to solve the gradient terms $\frac{\partial \boldsymbol{\phi}_{T+1}^{\text{BR}}}{\partial \vpi_T}, \frac{\partial \boldsymbol{\phi}_{T+1}^{\text{BR}}}{\partial \vPhi_{T}}$. 

Firstly, we can get an exact best-response by hundreds of gradient steps to achieve:
\begin{smalleralign}
\boldsymbol{\phi_{t+1}^{\text{BR}}} &= \mathop{\arg\max}_{\boldsymbol{\phi}} \mathfrak{M}\left(\boldsymbol{\phi}, \langle \vpi_{t}, \vPhi_{t} \rangle \right)
\end{smalleralign}

Since $\boldsymbol{\phi_{t+1}^{\text{BR}}}$ is a minimiser of the inner loop optimisation, we can derive the stationary point condition by implicit function theorem:
\begin{smalleralign}
& \frac{\partial \mathfrak{M}\left(\boldsymbol{\phi_{t+1}^{\text{BR}}}, \langle \vpi_{t}, \vPhi_{t} \rangle \right)}{\partial \boldsymbol{\phi_{t+1}^{\text{BR}}}} = 0\nonumber\\
& \rightarrow\frac{\partial}{\partial \vpi_{t}}\frac{\partial \mathfrak{M}\left(\boldsymbol{\phi_{t+1}^{\text{BR}}}, \langle \vpi_{t}, \vPhi_{t} \rangle \right)}{\partial \boldsymbol{\phi_{t+1}^{\text{BR}}}} =0 \nonumber\\
& \rightarrow\frac{\partial^2 \mathfrak{M}\left(\boldsymbol{\phi_{t+1}^{\text{BR}}}, \langle \vpi_{t}, \vPhi_{t} \rangle \right)}{\partial \boldsymbol{\phi_{t+1}^{\text{BR}}}{\partial \boldsymbol{\phi_{t+1}^{\text{BR}}}}^T} \frac{\partial \boldsymbol{\phi_{t+1}^{\text{BR}}}}{\partial \vpi_{t}} + \frac{\partial^{2} \mathfrak{M}\left(\boldsymbol{\phi_{t+1}^{\text{BR}}}, \langle \vpi_{t}, \vPhi_{t} \rangle \right)}{\partial \boldsymbol{\phi_{t+1}^{\text{BR}}} \partial \vpi_{t}} = 0 \nonumber\\
& \rightarrow\frac{\partial \boldsymbol{\phi_{t+1}^{\text{BR}}}}{\partial \vpi_t} = -\left[\frac{\partial^2 \mathfrak{M}\left(\boldsymbol{\phi_{t+1}^{\text{BR}}}, \langle \vpi_{t}, \vPhi_{t} \rangle \right)}{\partial \boldsymbol{\phi_{t+1}^{\text{BR}}}{\partial \boldsymbol{\phi_{t+1}^{\text{BR}}}}^T}\right]^{-1}\frac{\partial^{2} \mathfrak{M}\left(\boldsymbol{\phi_{t+1}^{\text{BR}}}, \langle \vpi_{t}, \vPhi_{t} \rangle \right)}{\partial \boldsymbol{\phi_{t+1}^{\text{BR}}} \partial \vpi_{t}}
\end{smalleralign}
Note that this implicit gradient requires the Hessian matrix $\frac{\partial^2 \mathfrak{M}(\boldsymbol{\phi}_{t+1}^{\text{BR}}, \langle\vpi_t, \boldsymbol{\Phi_{t}}\rangle)}{\partial \boldsymbol{\phi}_{t+1}^{\text{BR}}\partial {\boldsymbol{\phi}_{t+1}^{\text{BR}}}^{T}}$ to be invertible, so it may not hold in some situations (like normal form games). Following the same reasoning, we can get:
\begin{smalleralign}
\frac{\partial\boldsymbol{\phi}_{t+1}^{\text{BR}}}{\partial \vPhi_{t}}
&=  -\left[\frac{\partial^2 \mathfrak{M}(\boldsymbol{\phi}_{t+1}^{\text{BR}}, \langle\vpi_t, \boldsymbol{\Phi_{t}}\rangle)}{\partial \boldsymbol{\phi}_{t+1}^{\text{BR}}\partial {\boldsymbol{\phi}_{t+1}^{\text{BR}}}^{T}}\right]^{-1}\frac{\partial^{2} \mathfrak{M}(\boldsymbol{\phi}_{t+1}^{\text{BR}}, \langle \vpi_t, \boldsymbol{\Phi_{t}}\rangle)}{\partial \boldsymbol{\phi}_{t+1}^{\text{BR}} \partial \boldsymbol{\Phi_{t}}}.
\end{smalleralign}
\subsection{Reinforcement Learning Best-response Oracle with Direct Meta-gradient}
\label{pg_meta_gradient}
For a Reinforcement Learning based best-response oracle, the only difference is that we need to replace gradient terms with policy gradient estimation. We utilise first-order policy gradients for estimating $\frac{\partial \mathfrak{M}_{T+1}}{\partial \boldsymbol{\phi}_{T+1}^{\text{BR}}}, \frac{\partial \mathfrak{M}_{T+1}}{\partial \vpi_{T}}, \frac{\partial \mathfrak{M}_{T+1}}{\partial \vPhi_{T}}$. For the best-response process, a one-step RL example is to replace Eq. (\ref{ap:eq:one-step}) with:
\begin{smalleralign}
    \boldsymbol{\phi}_{t+1}^{\text{BR}}&=\boldsymbol{\phi}_{0}+\alpha \frac{\partial \mathfrak{M}\left(\boldsymbol{\phi}_0, \langle \vpi_t, \vPhi_{t} \rangle \right)}{\partial \boldsymbol{\phi}_0}\nonumber\\
    &=\boldsymbol{\phi}_{0}+\alpha\sum_{k=1}^{t}\vpi_{t}^{k} \nabla_{\boldsymbol{\phi}_{0}} E_{\boldsymbol{\tau}\sim P(\boldsymbol{\tau}|\boldsymbol{\phi}_{0}, \boldsymbol{\phi}_{k}^{\text{BR}})}\left[R(\tau)\right],
    \label{ap:eq:one-step-rl}
\end{smalleralign}

where $\alpha$ and $\boldsymbol{\tau}$ refer to the learning rate and the joint trajectories for two agents respectively. Reward $R$ represents the trajectory return for the first agent. The backward meta-gradient for the best-response process can be computed as:

\begin{smalleralign}
\frac{\partial \boldsymbol{\phi}_{t+1}^{\text{BR}}}{\partial \boldsymbol{\pi}_{t}}
&=\alpha \frac{\partial^{2} \mathfrak{M}\left(\boldsymbol{\phi}_{0},\left\langle\boldsymbol{\pi}_{t}, \boldsymbol{\Phi}_{t}\right\rangle\right)}{\partial \phi_{0} \partial \boldsymbol{\pi}_{t}}
\nonumber\\
&=\alpha \left(\nabla_{\boldsymbol{\phi}_{0}}E_{\boldsymbol{\tau}\sim P(\boldsymbol{\tau}|\boldsymbol{\phi}_{0}, \boldsymbol{\phi}_{k}^{\text{BR}})}\left[R(\tau)\right]\right)_{\{k=1,2,...,t\}},\\
\frac{\partial \boldsymbol{\phi}_{t+1}^{\text{BR}}}{\partial \boldsymbol{\Phi}_{t}}
&=\alpha \frac{\partial^{2} \mathfrak{M}\left(\phi_{0},\left\langle\boldsymbol{\pi}_{t}, \boldsymbol{\Phi}_{t}\right\rangle\right)}{\partial \phi_{0} \partial \boldsymbol{\Phi}_{t}}
\nonumber\\
&=\alpha \left(\nabla_{\boldsymbol{\phi}_{0}}\nabla_{\boldsymbol{\phi}_{k}^{\text{BR}}}E_{\boldsymbol{\tau}\sim P(\boldsymbol{\tau}|\boldsymbol{\phi}_{0}, \boldsymbol{\phi}_{k}^{\text{BR}})}\left[R(\tau)\right]\right)_{\{k=1,2,...,t\}}.
\end{smalleralign}

Eq. (\ref{eq:phi_theta}) for a Reinforcement Learning based oracle can be handled following a similar manner by replacing gradients with policy gradient estimation.

So the main issue is: how can we estimate the second-order policy gradient $\nabla_{\boldsymbol{\phi_{1}}}\nabla_{\boldsymbol{\phi_{2}}}E_{\tau \sim p(\tau|\boldsymbol{\phi_{1}},\boldsymbol{\phi_{2}})}[R(\boldsymbol{\tau})]$, where $\boldsymbol{\phi_1}$, $\boldsymbol{\phi_2}$ denotes policy for two agents. There are several higher order gradient estimators like DICE \cite{foerster2018dice}, LVC \cite{rothfuss2018promp} that can help us. In our case, we utilise DICE which is entirely compatible with automatic differentiation toolbox. In the following part, we follow similar analysis way like \cite{rothfuss2018promp} to show how second-order policy gradient is like and how we can estimate unbiased first-order and second-order policy gradient with DICE. In the following part, $P(\boldsymbol{\tau} \mid \boldsymbol{\phi_1}, \boldsymbol{\phi_2})$ and $P_{\boldsymbol{\phi_1}, \boldsymbol{\phi_2}}(\boldsymbol{\tau})$ represent the probability of the joint trajectory.

\begin{smalleralign}
&\nabla_{\boldsymbol{\phi_1}}\nabla_{\boldsymbol{\phi_2}} \mathbb{E}_{\boldsymbol{\tau} \sim P(\boldsymbol{\tau} \mid \boldsymbol{\phi_1}, \boldsymbol{\phi_2})}[R(\boldsymbol{\tau})]\nonumber\\&= \nabla_{\boldsymbol{\phi_1}} \mathbb{E}_{\boldsymbol{\tau} \sim P(\boldsymbol{\tau} \mid \boldsymbol{\phi_1}, \boldsymbol{\phi_2})}\left[\nabla_{\boldsymbol{\phi_2}} \log P_{\boldsymbol{\phi_1}, \boldsymbol{\phi_2}}(\boldsymbol{\tau}) R(\boldsymbol{\tau})\right] \nonumber\\
&= \nabla_{\boldsymbol{\phi_1}} \int P(\boldsymbol{\tau} \mid \boldsymbol{\phi_1}, \boldsymbol{\phi_2}) \nabla_{\boldsymbol{\phi_2}} \log P_{\boldsymbol{\phi_1}, \boldsymbol{\phi_2}}(\boldsymbol{\tau}) R(\boldsymbol{\tau}) d \boldsymbol{\tau} \nonumber\\
&= \int \Big[ P(\boldsymbol{\tau} \mid \boldsymbol{\phi_1}, \boldsymbol{\phi_2}) \nabla_{\boldsymbol{\phi_1}} \log P_{\boldsymbol{\phi_1}, \boldsymbol{\phi_2}}(\boldsymbol{\tau}) \nabla_{\boldsymbol{\phi_2}} \log P_{\boldsymbol{\phi_1}, \boldsymbol{\phi_2}}(\boldsymbol{\tau})^{\top} R(\boldsymbol{\tau})\nonumber\\
& \qquad \qquad \qquad \qquad  \qquad \qquad \qquad \qquad    + P(\boldsymbol{\tau} \mid \boldsymbol{\phi_1}, \boldsymbol{\phi_2}) \nabla_{\boldsymbol{\phi_1}}\nabla_{\boldsymbol{\phi_2}} \log P_{\boldsymbol{\phi_1}, \boldsymbol{\phi_2}}(\boldsymbol{\tau}) R(\boldsymbol{\tau}) \Big] d \boldsymbol{\tau} \nonumber\\
&= \mathbb{E}_{\boldsymbol{\tau} \sim P(\boldsymbol{\tau} \mid \boldsymbol{\phi_1}, \boldsymbol{\phi_2})}\left[R(\boldsymbol{\tau})\left(\nabla_{\boldsymbol{\phi_1}}\nabla_{\boldsymbol{\phi_2}} \log P_{\boldsymbol{\phi_1}, \boldsymbol{\phi_2}}(\boldsymbol{\tau})+\nabla_{\boldsymbol{\phi_1}} \log P_{\boldsymbol{\phi_1}, \boldsymbol{\phi_2}}(\boldsymbol{\tau}) \nabla_{\boldsymbol{\phi_2}} \log P_{\boldsymbol{\phi_1}, \boldsymbol{\phi_2}}(\boldsymbol{\tau})^{\top}\right)\right].
\end{smalleralign}
In fact, we can show that $\nabla_{\boldsymbol{\phi_1}}\nabla_{\boldsymbol{\phi_2}} \log P_{\boldsymbol{\phi_1}, \boldsymbol{\phi_2}}(\boldsymbol{\tau})=0$.
\begin{smalleralign}
\nabla_{\boldsymbol{\phi_1}} \nabla_{\boldsymbol{\phi_2}} \log P_{\boldsymbol{\phi_1}, \phi_{2}}(\tau)&=
\nabla_{\boldsymbol{\phi_1}} \nabla_{\boldsymbol{\phi_2}}\log \prod_{i=0}^{n}P_{\boldsymbol{\phi_1},\boldsymbol{\phi_2}}(a_i^1,a_i^2|s_i^1,s_i^2)\nonumber\\
&=\nabla_{\boldsymbol{\phi_1}} \nabla_{\boldsymbol{\phi_2}}\log \prod_{i=0}^{n}\pi_{\boldsymbol{\phi_1}}(a_i^1|s_i^1)\pi_{\boldsymbol{\phi_2}}(a_i^2|s_i^2)\nonumber\\
&=\nabla_{\phi_{1}} \nabla_{\boldsymbol{\phi_2}}\sum_{i=0}^n \left(\log(\pi_{\boldsymbol{\phi_1}}(a_i^1|s_i^1))+\log(\pi_{\boldsymbol{\phi_2}}(a_i^2|s_i^2))\right)\nonumber\\
&=0
\end{smalleralign}
where $n$ denotes the length of the RL trajectory, $\vpi_{\boldsymbol{\phi_1}}$ and $\pi_{\boldsymbol{\phi_2}}$ represent stochastic policies for two agents respectively. Note that $P_{\boldsymbol{\phi_1},\boldsymbol{\phi_2}}(a_i^1,a_i^2|s_i^1,s_i^2) = \pi_{\boldsymbol{\phi_1}}(a_i^1|s_i^1)\pi_{\boldsymbol{\phi_2}}(a_i^2|s_i^2)$ because the agent only relies on its own state. Following \cite{foerster2018dice}'s formulation, we have:
\begin{smalleralign}
J^{\mathrm{DICE}} &=\sum_{t=0}^{H-1} \square\left(\left\{a_{j \in\{1,2\}}^{t^{\prime} \leq t}\right\}\right) r_t \nonumber\\
&=\sum_{t=0}^{H-1} \exp \left(\sum_{t^{\prime}=0}^{t} \log \pi_{\boldsymbol{\phi_1}}\left(a_{t^{\prime}}^{1} \mid s_{t^{\prime}}^{1}\right)\log \pi_{\boldsymbol{\phi_2}}\left(a_{t^{\prime}}^{2} \mid s_{t^{\prime}}^{2}\right)
-\perp \left(\log \pi_{\boldsymbol{\phi_1}}(a_{t^{\prime}}^{1} \mid s_{t^{\prime}}^{1}) \log \pi_{\boldsymbol{\phi_1}}(a_{t^{ \prime}}^{2} \mid s_{t^{}}^{2})\right)\right) r_{t}
\end{smalleralign}

We denote $\perp$ as the "stop gradient" operator and $\rightarrow$ as the "evaluates to" operator. "Evaluates to" operator $\rightarrow$ is in contrast with =, which also brings the equality of gradients. So the "stop gradient" operator here means that $\perp\left(f_{\vtheta}(x)\right) \rightarrow f_{\vtheta}(x)$ but $\nabla_{\vtheta} \perp\left(f_{\vtheta}(x)\right) \rightarrow 0$.

To make the DICE loss concise, we reformulate it as follows:
\begin{equation}
J^{\mathrm{DICE}}=\sum_{t=0}^{H-1}\left(\prod_{t^{\prime}=0}^{t} \frac{\pi_{\boldsymbol{\phi_1}}\left(a_{t^{\prime}}^{1} \mid s_{t^{\prime}}^{1} \right)\pi_{\boldsymbol{\phi_2}}(a_{t^{\prime}} ^{2}\mid s_{t^{\prime}}^{2})}{\perp\left(\pi_{\boldsymbol{\phi_1}}\left(a_{t^{\prime}} ^{1}\mid s_{t^{\prime}}^{1}\right)\pi_{\boldsymbol{\phi_2}}\left(a_{t^{\prime}} ^{2}\mid s_{t^{\prime}}^{2}\right)\right)}\right) r_t
\end{equation}

where $r_t$ refers to the reward agent 1 gets at timestep $t$.
\begin{smalleralign}
\nabla_{\boldsymbol{\phi_1}} J^{\mathrm{DICE}} &=\sum_{t=0}^{H-1} \nabla_{\boldsymbol{\phi_1}}\left(\prod_{t^{\prime}=0}^{t} \frac{\pi_{\boldsymbol{\phi_1}}\left(a_{t^{\prime}}^{1} \mid s_{t^{\prime}}^{1} \right)\pi_{\boldsymbol{\phi_2}}(a_{t^{\prime}} ^{2}\mid s_{t^{\prime}}^{2})}{\perp\left(\pi_{\boldsymbol{\phi_1}}\left(a_{t^{\prime}} ^{1}\mid s_{t^{\prime}}^{1}\right)\pi_{\boldsymbol{\phi_2}}\left(a_{t^{\prime}} ^{2}\mid s_{t^{\prime}}^{2}\right)\right)}\right) r_t\nonumber\\
&=\sum_{t=0}^{H-1}\left(\prod_{t^{\prime}=0}^{t} \frac{\pi_{\boldsymbol{\phi_1}}\left(a_{t^{\prime}}^{1} \mid s_{t^{\prime}}^{1} \right)\pi_{\boldsymbol{\phi_2}}(a_{t^{\prime}} ^{2}\mid s_{t^{\prime}}^{2})}{\perp\left(\pi_{\boldsymbol{\phi_1}}\left(a_{t^{\prime}} ^{1}\mid s_{t^{\prime}}^{1}\right)\pi_{\boldsymbol{\phi_2}}\left(a_{t^{\prime}} ^{2}\mid s_{t^{\prime}}^{2}\right)\right)}\right)\left(\sum_{t^{\prime}=0}^{t} \nabla_{\boldsymbol{\phi_1}} \log \pi_{\boldsymbol{\phi_1}}\left(a_{t^{\prime}}^1 \mid s_{t^{\prime}}^1\right)\right) r_t \nonumber\\
& \rightarrow \sum_{t=0}^{H-1}\left(\sum_{t^{\prime}=0}^{t} \nabla_{\boldsymbol{\phi_{1}}} \log \pi_{\boldsymbol{\phi_1}}\left(a_{t^{\prime}}^1 \mid s_{t^{\prime}}^1\right)\right) r_t
\end{smalleralign}

\begin{smalleralign}
\mathbb{E}_{\boldsymbol{\tau} \sim P(\boldsymbol{\tau} \mid \boldsymbol{\phi_1}, \boldsymbol{\phi_2})}\left[\nabla_{\boldsymbol{\phi_1}} J^{\mathrm{DICE}}\right] &=\mathbb{E}_{\boldsymbol{\tau} \sim P(\boldsymbol{\tau} \mid \boldsymbol{\phi_1}, \boldsymbol{\phi_2})}\left[\sum_{t=0}^{H-1}\left(\sum_{t^{\prime}=0}^{t} \nabla_{\boldsymbol{\phi_{1}}} \log \pi_{\boldsymbol{\phi_1}}\left(a_{t^{\prime}}^1 \mid s_{t^{\prime}}^1\right)\right) r_t\right]\nonumber\nonumber\\
&=\nabla_{\boldsymbol{\phi_1}}\mathbb{E}_{\boldsymbol{\tau} \sim P(\boldsymbol{\tau} \mid \boldsymbol{\phi_1}, \boldsymbol{\phi_2})}\left[R(\boldsymbol{\tau})\right]
\end{smalleralign}
which corresponds to standard policy gradients for single-agent in a multi-agent environment (agents will consider other agents as part of the environment). And the hessian for the DICE loss is:
\begin{smalleralign}
&\nabla_{\boldsymbol{\phi_2}}\nabla_{\boldsymbol{\phi_1}} J^{\mathrm{DICE}}\nonumber\\
&=\sum_{t=0}^{H-1} \nabla_{\boldsymbol{\phi_2}}\left(\prod_{t^{\prime}=0}^{t} \frac{\pi_{\boldsymbol{\phi_1}}\left(a_{t^{\prime}}^{1} \mid s_{t^{\prime}}^{1}\right) \pi_{\boldsymbol{\phi_2}}\left(a_{t^{\prime}}^{2} \mid s_{t^{\prime}}^{2}\right)}{\perp\left(\pi_{\boldsymbol{\phi_1}}\left(a_{t^{\prime}}^{1} \mid s_{t^{\prime}}^{1}\right) \pi_{\boldsymbol{\phi_2}}\left(a_{t^{\prime}}^{2} \mid s_{t^{\prime}}^{2}\right)\right)}\right) \left(\sum_{t^{\prime}=0}^{t} \nabla_{\boldsymbol{\phi_1}} \log \pi_{\boldsymbol{\phi_1}}\left(a_{t^{\prime}}^{1} 
\mid s_{t^{\prime}}^{1}\right)\right) r_{t} \nonumber\\
&=\sum_{t=0}^{H-1}\left(\prod_{t^{\prime}=0}^{t} \frac{\pi_{\boldsymbol{\phi_1}}\left(a_{t^{\prime}}^{1} \mid s_{t^{\prime}}^{1}\right) \pi_{\boldsymbol{\phi_2}}\left(a_{t^{\prime}}^{2} \mid s_{t^{\prime}}^{2}\right)}{\perp\left(\pi_{\boldsymbol{\phi_1}}\left(a_{t^{\prime}}^{1} \mid s_{t^{\prime}}^{1}\right) \pi_{\boldsymbol{\phi_2}}\left(a_{t^{\prime}}^{2} \mid s_{t^{\prime}}^{2}\right)\right)}\right)\cdot  \nonumber \\  
& \qquad \qquad \qquad \qquad  \qquad \qquad \qquad   \left(\sum_{t^{\prime}=0}^{t} \nabla_{\boldsymbol{\phi_1}} \log \pi_{\boldsymbol{\phi_1}}\left(a_{t^{\prime}}^{1} \mid s_{t^{\prime}}^{1}\right)\right)  \left(\sum_{t^{\prime}=0}^{t} \nabla_{\boldsymbol{\phi_2}} \log \pi_{\boldsymbol{\phi_2}}\left(a_{t^{\prime}}^{2} \mid s_{t^{\prime}}^{2}\right)\right)^{\top}r_{t} \nonumber\\
& \text{which can be evaluated via the following: } \nonumber \\ 
& \rightarrow \sum_{t=0}^{H-1}\left(\sum_{t^{\prime}=0}^{t} \nabla_{\boldsymbol{\phi_{1}}} \log \pi_{\boldsymbol{\phi_1}}\left(a_{t^{\prime}}^{1} \mid s_{t^{\prime}}^{1}\right)\right) \left(\sum_{t^{\prime}=0}^{t} \nabla_{\boldsymbol{\phi_{2}}} \log \pi_{\boldsymbol{\phi_2}}\left(a_{t^{\prime}}^{2} \mid s_{t^{\prime}}^{2}\right)\right)^{\top}r_{t}
\end{smalleralign}
So finally we have:
\begin{smalleralign}
&\mathbb{E}_{\boldsymbol{\tau} \sim P_{\mathcal{T}}(\boldsymbol{\tau} \mid \boldsymbol{\phi_1}, \boldsymbol{\phi_2})}\left[ \nabla_{\boldsymbol{\phi_{1}}} \nabla_{\boldsymbol{\phi_{2}}} J^{\mathrm{DICE}}\right]\nonumber\\ &=\mathbb{E}_{\boldsymbol{\tau} \sim P\left(\boldsymbol{\tau} \mid \boldsymbol{\phi_1}, \boldsymbol{\phi_2}\right)}\left[\sum_{t=0}^{H-1}\left(\sum_{t^{\prime}=0}^{t} \nabla_{\boldsymbol{\phi_{1}}} \log \pi_{\boldsymbol{\phi_{1}}}\left(a_{t^{\prime}}^{1} \mid s_{t^{\prime}}^{1}\right)\right)\left(\sum_{t^{\prime}=0}^{t} \nabla_{\boldsymbol{\phi_{2}}} \log \pi_{\boldsymbol{\phi_{2}}}\left(a_{t^{\prime}}^{2} \mid s_{t^{\prime}}^{2}\right)\right)^{\top} r_{t}\right]\nonumber\\
&= \mathbb{E}_{\boldsymbol{\tau} \sim P\left(\boldsymbol{\tau} \mid\boldsymbol{\phi_1}, \boldsymbol{\phi_2}\right)}\left[R(\boldsymbol{\tau})\nabla_{\boldsymbol{\phi_{1}}} \log P_{\boldsymbol{\phi_{1}, \phi_{2}}}(\boldsymbol{\tau}) \nabla_{\boldsymbol{\phi_{2}}} \log P_{\boldsymbol{\phi_{1}}, \boldsymbol{\phi_{2}}}(\boldsymbol{\tau})^{\top}\right]\nonumber\\
&= \nabla_{\boldsymbol{\phi_{1}}} \nabla_{\boldsymbol{\phi_{2}}} \mathbb{E}_{\boldsymbol{\tau} \sim P\left(\boldsymbol{\tau} \mid \boldsymbol{\phi_{1}}, \boldsymbol{\phi_{2}}\right)}[R(\boldsymbol{\tau})]
\end{smalleralign}

In all, we have shown that by plugging DICE into the computation graph, we can obtain unbiased first-order and second-order policy gradient estimation and also the overall meta-gradient estimation.
\newpage
\section{Pseudo-Codes of the Proposed Algorithms}
\label{pseudo}
\subsection{Gradient-Descent Best-Response  Oracles}
\subsubsection*{Non-Implicit Version}
Here we provide details of NAC where few-step gradient descent is used as the best-response oracle in Alg. (\ref{alg:auto-psro-gd-nonimp}), and therefore we are in the non-implicit setting.
\begin{algorithm}[H]
\caption{NAC with Gradient-Descent Best-Response  Oracles}
\label{alg:auto-psro-gd-nonimp}
\begin{algorithmic}[1]
\Require Game distribution $p(\mG)$, inner learning rate $\alpha$, outer learning rate $\beta$, time window $T$.
\State Randomly initialise policy pool $\boldsymbol{\phi_{0}}$, Initialise parameters $\vtheta$ of the meta solver $f_\vtheta$.
\For{each training iteration}
\State Sample games $\{G_{k}\}_{k=1,...,K}$ from $p(\mG)$. 
\For{each game $G_k$}
\For{each  iteration $t$}
\State Compute the meta-policy   $\vpi_{t-1}=f(\tM_{t-1})$. 
\State Initialise random best-response policy $\boldsymbol{\phi}_0$.
\For{gradient updates $n$}
\State Compute $\boldsymbol{\phi}_{n+1}=\boldsymbol{\phi}_{n}+\alpha \frac{\partial \mathfrak{M}\left(\boldsymbol{\phi}_n, \langle \vpi_{t-1}, \vPhi_{t-1} \rangle \right)}{\partial \boldsymbol{\phi}_n}$ via Eq. (\ref{eq:one-step})
\EndFor
\State Expand the population $\boldsymbol{\Phi_{t}} = \boldsymbol{\Phi_{t-1}} \cup \{\boldsymbol{\phi}_{t}^{\text{BR}}\}$ 
\EndFor
\State Compute the meta-policy   $\vpi_{T}=f(\tM_{T})$. 
\State Compute $\mathfrak{Exp}_i(\vpi_T, \boldsymbol{\Phi}_T)$ by Eq. (\ref{exp})
\EndFor
\State{Compute the meta-gradient $\textbf{g}_k$ via Eq. (\ref{meta_gradient})} with br meta-gradient following Eq. (\ref{eq:br_bp})
\State Update meta-solver's parameters $\vtheta^{\prime} = \vtheta - \beta \frac{1}{K} \sum_k{\textbf{g}_k}$.
\EndFor
\end{algorithmic}
\end{algorithm}

\subsubsection*{Implicit Version}
Here we provide details of NAC where many-step gradient descent is used as the best-response oracle in Alg. (\ref{alg:auto-psro-gd-imp}), and therefore we are in the implicit setting.
\begin{algorithm}[H]
\caption{NAC with Gradient-Descent Best-Response  Oracles-Implicit}
\label{alg:auto-psro-gd-imp}
\begin{algorithmic}[1]
\Require Game distribution $p(\mG)$, inner learning rate $\alpha$, outer learning rate $\beta$, time window $T$.
\State Randomly initialise policy pool $\boldsymbol{\phi_{0}}$, Initialise parameters $\vtheta$ of the meta solver $f_\vtheta$.
\For{each training iteration}
\State Sample games $\{G_{k}\}_{k=1,...,K}$ from $p(\mG)$. 
\For{each game $G_k$}
\For{each  iteration $t$}
\State Compute the meta-policy   $\vpi_{t-1}=f(\tM_{t-1})$. 
\State Initialise random best-response policy $\boldsymbol{\phi}_0$.
\For{gradient updates $n$ (Large $n$)}
\State Compute $\boldsymbol{\phi}_{n+1}^{\text{BR}}=\boldsymbol{\phi}_{n}+\alpha \frac{\partial \mathfrak{M}\left(\boldsymbol{\phi}_n, \langle \vpi_{t-1}, \vPhi_{t-1} \rangle \right)}{\partial \boldsymbol{\phi}_n}$ via Eq. (\ref{eq:one-step})
\EndFor
\State Expand the population $\boldsymbol{\Phi_{t}} = \boldsymbol{\Phi_{t-1}} \cup \{\boldsymbol{\phi}_{t}^{\text{BR}}\}$ 
\EndFor
\State Compute the meta-policy   $\vpi_{T}=f(\tM_{T})$. 
\State Compute $\mathfrak{Exp}_i(\vpi_T, \boldsymbol{\Phi}_T)$ by Eq. (\ref{exp})
\EndFor
\State{Compute the meta-gradient $\textbf{g}_k$ via Eq. (\ref{meta_gradient})} with br meta-gradient following Eq. (\ref{eq:imp2})
\State Update meta-solver's parameters $\vtheta^{\prime} = \vtheta - \beta \frac{1}{K} \sum_k{\textbf{g}_k}$.
\EndFor
\end{algorithmic}
\end{algorithm}

\subsection{Reinforcement Learning Best-Response Oracles}
Here we provide details of NAC where reinforcement learning is used as the best-response oracle in Alg. (\ref{alg:auto-psro-rl}), where we apply DICE for unbiased meta-gradient estimation.
\begin{algorithm}[H]
\caption{NAC with Reinforcement Learning Best-Response Oracles}
\label{alg:auto-psro-rl}
\begin{algorithmic}[1]
\Require Game distribution $p(\mG)$, inner learning rate $\alpha$, outer learning rate $\beta$, time window $T$.
\State Randomly initialise policy pool $\boldsymbol{\phi_{0}}$, Initialise parameters $\vtheta$ of the meta solver $f_\vtheta$.
\For{each training iteration}
\State Sample games $\{G_{k}\}_{k=1,...,K}$ from $p(\mG)$. 
\For{each game $G_k$}
\For{each  iteration $t$}
\State Compute the meta-policy   $\vpi_{t-1}=f(\tM_{t-1})$. 
\State Initialise random best-response policy $\boldsymbol{\phi}_0$.
\For{gradient updates $n$}
\State Compute $\boldsymbol{\phi}_{n+1}=\boldsymbol{\phi}_{n}+\alpha \frac{\partial \mathfrak{M}\left(\boldsymbol{\phi}_n, \langle \vpi_{t-1}, \vPhi_{t-1} \rangle \right)}{\partial \boldsymbol{\phi}_n}$ with DICE in Eq. (\ref{eq:dice})
\EndFor
\State Expand the population $\boldsymbol{\Phi_{t}} = \boldsymbol{\Phi_{t-1}} \cup \{\boldsymbol{\phi}_{t}^{\text{BR}}\}$ 
\EndFor
\State Compute the meta-policy   $\vpi_{T}=f(\tM_{T})$. 
\State Compute $\mathfrak{Exp}_i(\vpi_T, \boldsymbol{\Phi}_T)$ by Eq. (\ref{exp})
\EndFor
\State{Compute the meta-gradient $\textbf{g}_k$ via Eq. (\ref{meta_gradient})} obtained by differentiating DICE in Eq. (\ref{eq:dice})
\State Update meta-solver's parameters $\vtheta^{\prime} = \vtheta - \beta \frac{1}{K} \sum_k{\textbf{g}_k}$.
\EndFor
\end{algorithmic}
\end{algorithm}

\newpage
\subsection{Optimising the Meta-Solver through Evolution Strategies}
\subsubsection*{ES-NAC with Approximate Tabular Best-response V1}
Here we provide details of NAC-ES where we use Tabular Approximate Best-Response V1 as the best-response oracle in Alg. (\ref{alg:tab-v1}).
\begin{algorithm}[H]
\caption{ES-NAC with Approximate Tabular Best-response V1}
\label{alg:tab-v1}
\begin{algorithmic}[1]
\Require Game distribution $p(\mG)$, outer learning rate $\alpha$, time window $T$, perturbations $n$, precision $\sigma$.
\State Randomly initialise policy pool $\boldsymbol{\phi_{0}}$, Initialise parameters $\vtheta$ of the meta solver $f_\vtheta$.
\For{each training iteration}
\State Sample games $\{G_{k}\}_{k=1,...,K}$ from $p(\mG)$. \State Sample $\boldsymbol{\epsilon}_1,...,\boldsymbol{\epsilon}_n \sim \mathcal{N}(0, I)$ and store $n$ models $f_{(\vtheta + \boldsymbol{\epsilon}_i)}$.
\For{each stored model $f$}
\For{each game $G_k$}
\For{each  iteration $t$}
\State Compute the meta-policy   $\vpi_{t-1}=f(\tM_{t-1})$. 
\State Initialise tabular best-response policy $\boldsymbol{\phi}^{\text{BR}}_{t}$.
\For{each state $s$}
\For{each action $a$}
\State Get exp. val. of $a$ against $\vpi_{t-1}$, $\mathbb{E}_{\vpi_{t-1}}(v(a)|s)$, by traversing game-tree.
\EndFor
\State Compute $a' = \argmax_a \mathbb{E}_{\vpi_{t-1}}(v(a)|s)$
\State Set $\boldsymbol{\phi}_{t}^{\text{BR}}(a'|s) = 0.75$ and $\boldsymbol{\phi}_{t}^{\text{BR}}(a^{\neg\prime}|s) = 0.25$
\EndFor
\State Expand the population $\boldsymbol{\Phi_{t}} = \boldsymbol{\Phi_{t-1}} \cup \{\boldsymbol{\phi}_{t}^{\text{BR}}\}$ 
\EndFor
\State Compute the meta-policy   $\vpi_{T}=f(\tM_{T})$. 
\State Compute $\mathfrak{Exp}_i(\vpi_T, \boldsymbol{\Phi}_T)$ by Eq. (\ref{exp})
\EndFor
\EndFor
\State{Compute the meta-gradient $\textbf{g}_k$ via Eq. (\ref{es:metagrad})} 
\State Update meta-solver's parameters $\vtheta^{\prime} = \vtheta - \alpha \frac{1}{K} \sum_k{\textbf{g}_k}$.
\EndFor
\end{algorithmic}
\end{algorithm}
\newpage
\subsubsection*{ES-NAC with Approximate Tabular Best-response V2}
Here we provide details of NAC-ES where we use Tabular Approximate Best-Response V2 as the best-response oracle in Alg. (\ref{alg:tab-v2}).
\begin{algorithm}[H]
\caption{ES-NAC with Approximate Tabular Best-response V2}
\label{alg:tab-v2}
\begin{algorithmic}[1]
\Require Game distribution $p(\mG)$, outer learning rate $\alpha$, time window $T$, perturbations $n$, precision $\sigma$.
\State Randomly initialise policy pool $\boldsymbol{\phi_{0}}$, Initialise parameters $\vtheta$ of the meta solver $f_\vtheta$.
\For{each training iteration}
\State Sample games $\{G_{k}\}_{k=1,...,K}$ from $p(\mG)$. 
\State Sample $\boldsymbol{\epsilon}_1,...,\boldsymbol{\epsilon}_n \sim \mathcal{N}(0, I)$ and store $n$ models $f_{(\vtheta + \boldsymbol{\epsilon}_i)}$.
\For{each stored model $f$}
\For{each game $G_k$}
\For{each  iteration $t$}
\State Compute the meta-policy   $\vpi_{t-1}=f(\tM_{t-1})$. 
\State Initialise tabular best-response policy $\boldsymbol{\phi}^{\text{BR}}_{t}$.
\For{each state $s$}
\For{each action $a$}
\State Get exp. val. of $a$ against $\vpi_t$, $\mathbb{E}_{\vpi_{t-1}}(v(a)|s)$, by traversing game-tree.
\EndFor
\State Compute $a' = \argmax_a \mathbb{E}_{\vpi_{t-1}}(v(a)|s)$
\State Sample $\boldsymbol{\eta}_1,\boldsymbol{\eta}_2 \sim \mathcal{N}(0, 1)$
\State Set $\boldsymbol{\phi}_{t}^{\text{BR}}(a'|s) = 1 + \boldsymbol{\eta}_1$ and $\boldsymbol{\phi}_{t}^{\text{BR}}(a^{\neg\prime}|s) = \boldsymbol{\eta}_2$
\State Normalise $\boldsymbol{\phi}_{t}^{\text{BR}}(s)$
\EndFor
\State Expand the population $\boldsymbol{\Phi_{t}} = \boldsymbol{\Phi_{t-1}} \cup \{\boldsymbol{\phi}_{t}^{\text{BR}}\}$ 
\EndFor
\State Compute the meta-policy   $\vpi_{T}=f(\tM_{T})$. 
\State Compute $\mathfrak{Exp}_i(\vpi_T, \boldsymbol{\Phi}_T)$ by Eq. (\ref{exp})
\EndFor
\EndFor
\State{Compute the meta-gradient $\textbf{g}_k$ via Eq. (\ref{es:metagrad})} 
\State Update meta-solver's parameters $\vtheta^{\prime} = \vtheta - \alpha \frac{1}{K} \sum_k{\textbf{g}_k}$.
\EndFor
\end{algorithmic}
\end{algorithm}
\newpage
\section{Additional Experimental Results}
The Pseudo code is in Appendix \ref{pseudo}. In this section we offer additional experimental results for better illustration of NAC.
\subsection{Kuhn Poker Experiments}
\label{ap:kuhn-poker}
\begin{figure*}[htbp]
 		\centering
        \includegraphics[width=.65\linewidth]{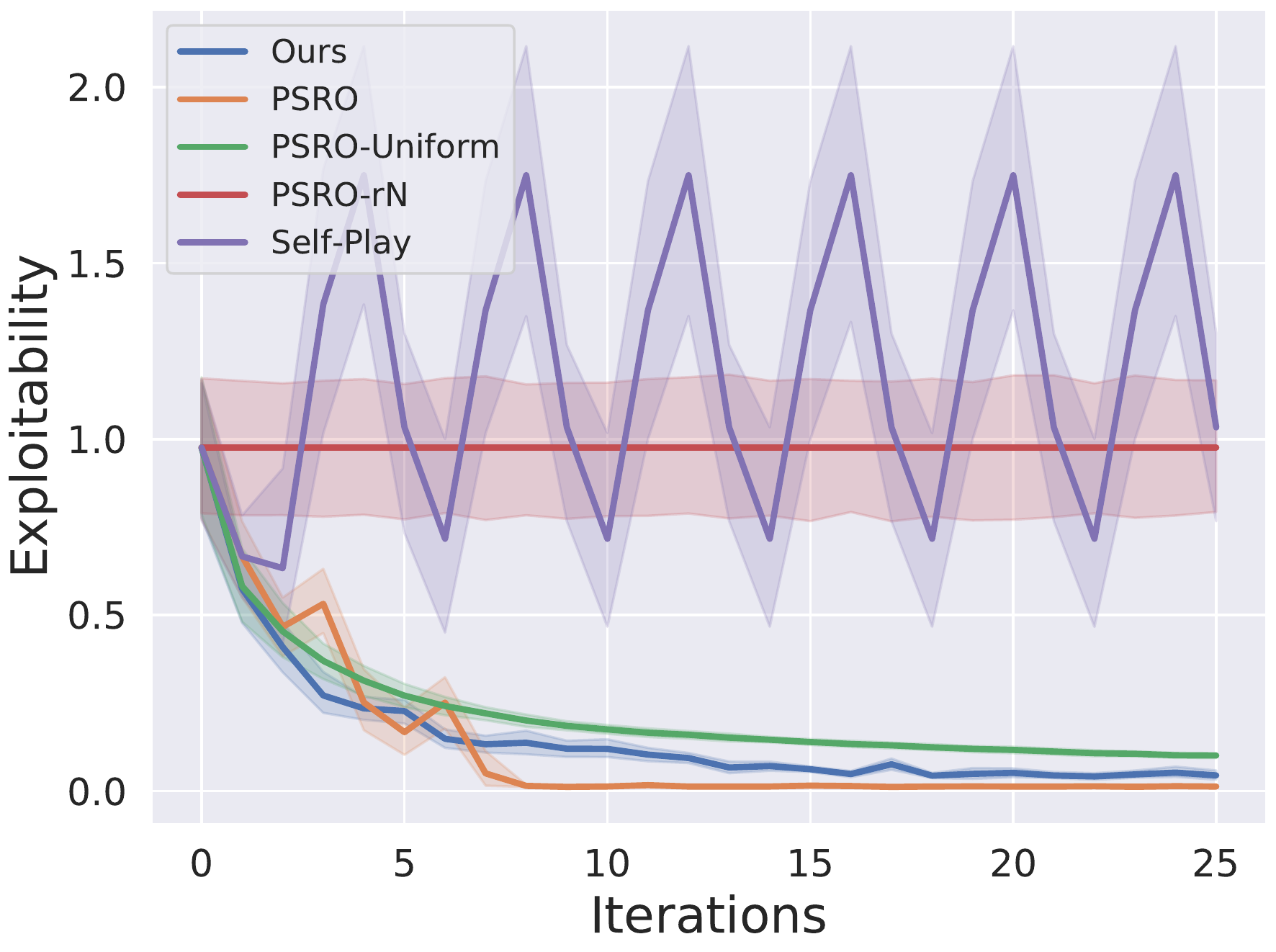}
        \caption{In-task training performance on Kuhn Poker using an exact tabular best-response oracle.}
 		\label{ap:fig:kpexact}
 	\end{figure*}

We provide the in-task training results for the Kuhn Poker exact tabular best-response method in Fig. (\ref{ap:fig:kpexact}), which was used to generate the exact best-response generalisation results from Kuhn Poker to Leduc Poker in Fig. (\ref{fig:gen}). Notably, whilst our model is slightly outperformed by PSRO, both achieve an exploitability of very close to 0 and therefore have both converged to an $\epsilon$-Nash equilibrium.

\newpage
\subsection{Visualisation of the Learned Curricula}
\label{visualisation}
\begin{figure*}[htbp]
 		\centering
        \includegraphics[width=\linewidth]{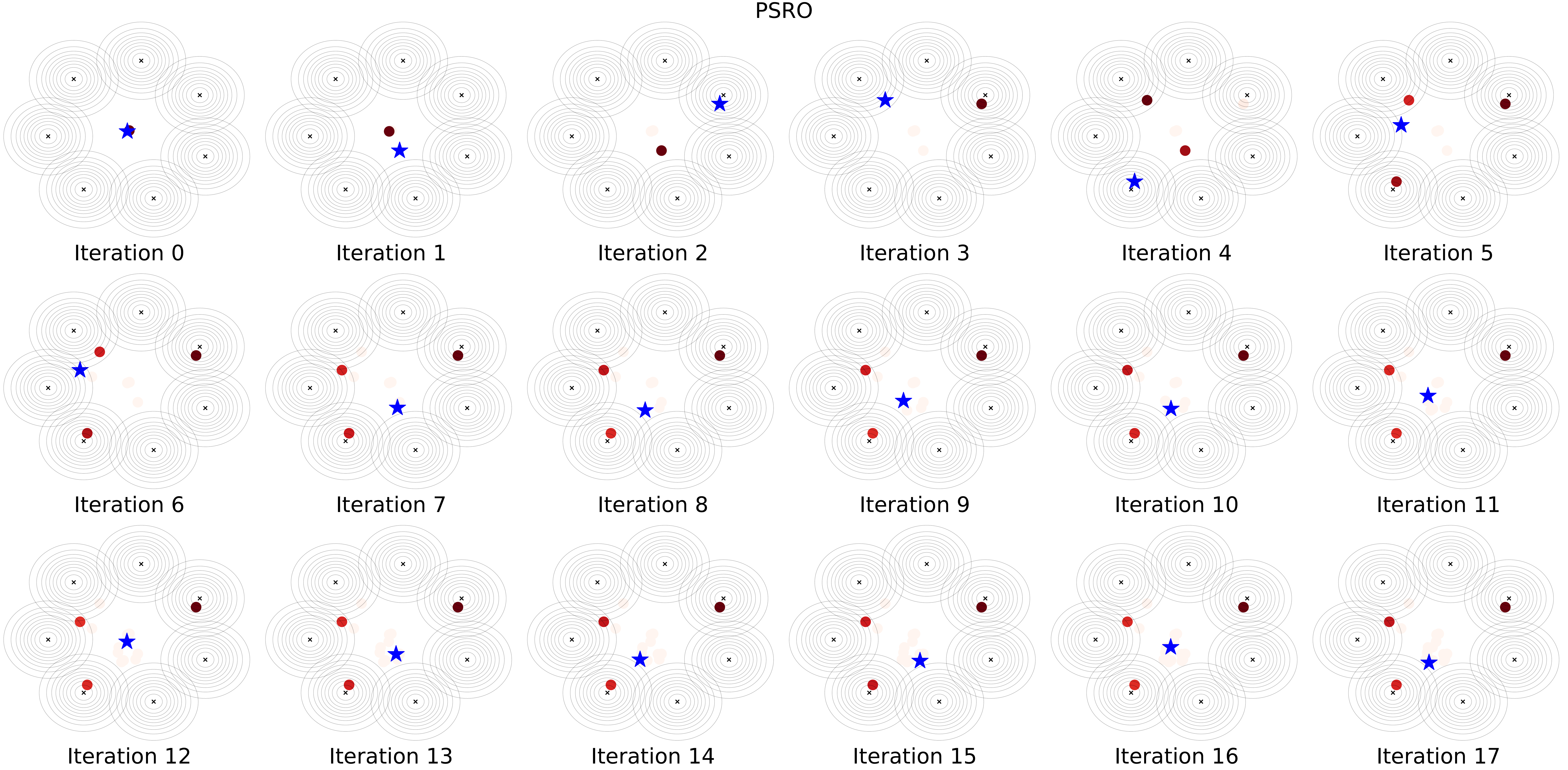}
        \caption{Visualisation of the full curricula on 2D-RPS using PSRO. Red points denote the meta-solver output, and darker refers to higher probability in $\vpi$. The blue star is the latest best-response.}
 		\label{ap:fig:psro_all_visualisation}
 	\end{figure*}

\begin{figure*}[htbp]
 		\centering
        \includegraphics[width=\linewidth]{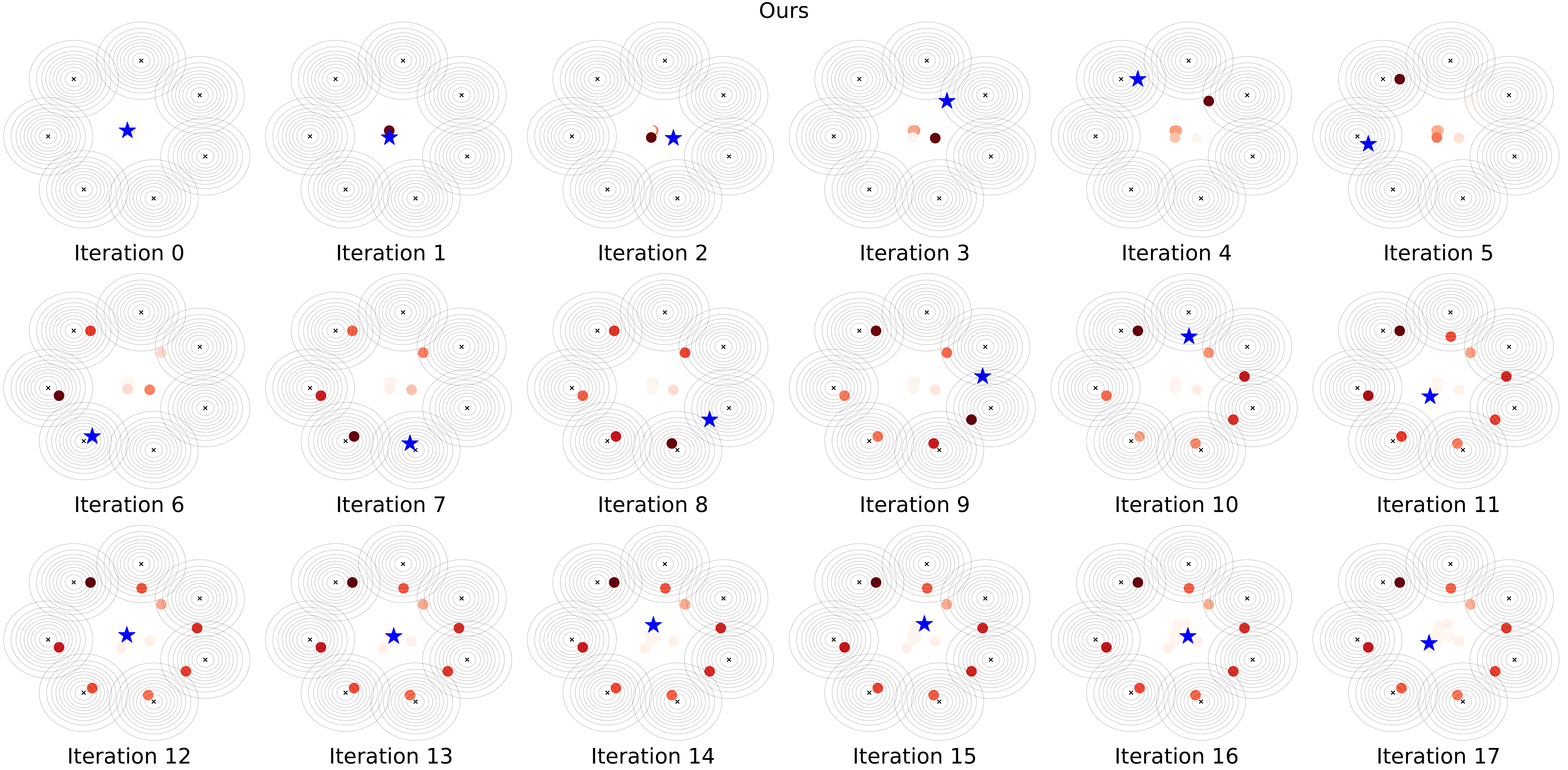}
        \caption{Visualisation of the full curricula on 2D-RPS using NAC. Red points denote the meta-solver output, and darker refers to higher probability in $\vpi$. The blue star is the latest best-response.}
 		\label{ap:fig:NAC_all_visualisation}
 	\end{figure*}

In Fig. (\ref{visualisation}) we offer a truncated view into the auto-curricula generated by PSRO and NAC. Here, we extend the visualisation to the full iterative process for PSRO and NAC in Fig. (\ref{ap:fig:psro_all_visualisation}) and Fig. (\ref{ap:fig:NAC_all_visualisation}) respectively. Due to the the approximate best-response setting, PSRO converges at iteration 7 and fail to reach all 7 Gaussian distributions. We suspect this is because fictitious play can only generate a difficult to beat auto-curricula without considering whether the best-response process is capable of learning a strong enough policy. In contrast, NAC generates a more smooth and appropriate auto-curricula, in which even an approximate best-response is able to learn a useful policy to explore each distribution one by one. Finally NAC essentially explores all 7 Gaussian distributions and achieves lower exploitability. Another interesting point is that the meta-solver only offers higher probability over the points near the Gaussian centres, which validates its ability to accurately evaluate the policies in a population.

 	\vspace{-15pt}
 \clearpage
 
In addition, we offer a similar visualisation of 12 iterations' worth of policy distributions produced by NAC on Kuhn Poker. Due to the approximate best-response limitation, it's difficult for NAC to get the exact Nash equilibrium policy. However, the final policy distribution of NAC still achieves a great approximation to the exact one. 	
 	\begin{figure*}[htbp]
 		\centering
        \includegraphics[width=\linewidth]{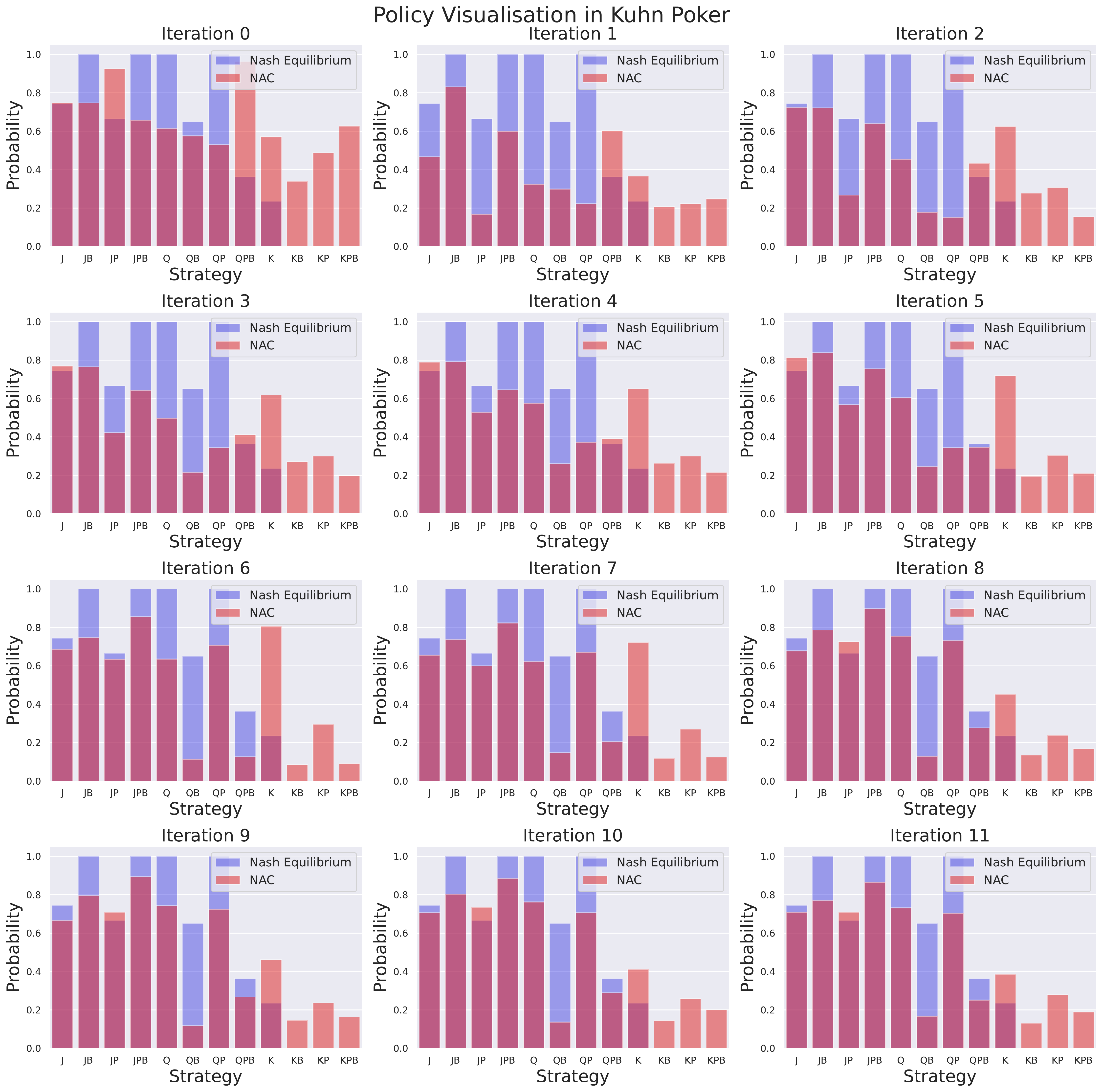}
        \caption{Visualisations of the whole 12 iterations policy distribution on Kuhn Poker for NAC and exact Nash Equilibrium. The Orange Red bar and Lighr blue bar refer to policy distribution for NAC and exact Nash Equilibrium respectively. The pink red bar represents the overlapping policy distribution.}
 		\label{ap:fig:kuhn_all_visualisation}
 	\end{figure*}
\clearpage
 
\subsection{Meta-Game Generalisation Results}
\label{ap:gener}
\cite{czarnecki2020real} introduced the concept of Games of Skill where certain real-world games share a similar structure in terms of their respective meta-games, and this work additionally released a collection of meta-games sharing this structure. As we utilise Randomly generated Games of Skill as our training game for our NFG meta-solver, we additionally test the ability of our learned meta-solver to generalise to unseen Games of Skill in the collection of meta-games from \cite{czarnecki2020real}.
\begin{figure}[H]
    \centering
    \includegraphics[width=.85\textwidth]{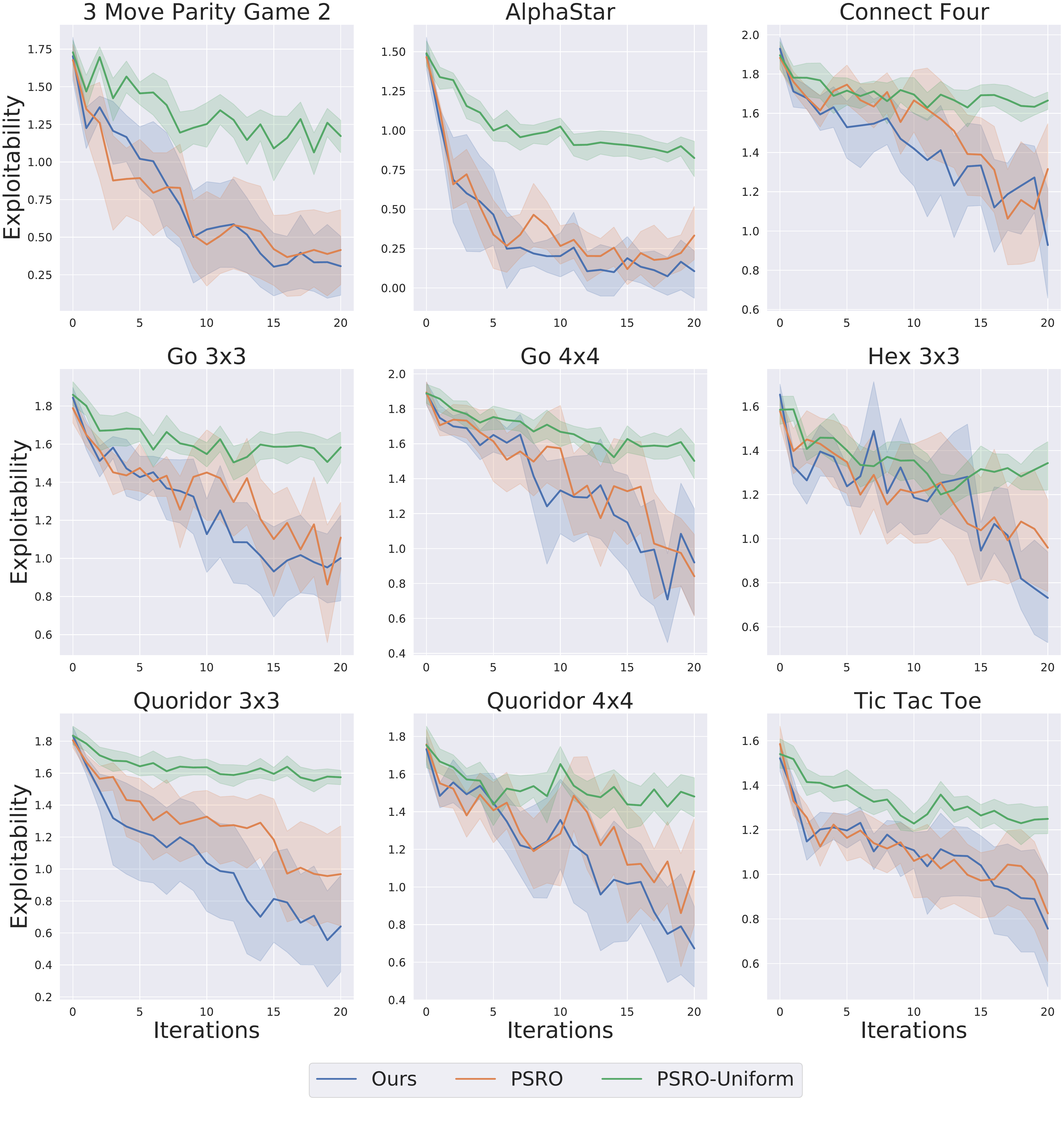}
    \caption{Exploitability results on the meta-games introduced in \cite{czarnecki2020real}. NAC performs at least as well as the best baseline in all settings, and often outperforms the PSRO baselines.}
    \label{ap:spgen}
\end{figure}
\newpage
\subsection{Meta-solver trained with Reinforcement Learning}
\label{ap:train_rl}

In our paper, we also train the meta-solver with Reinforcement Learning. Note that RL here refers to the technique for training the meta-solver rather than the best-response oracle.  In particular, 
we can treat the whole PSRO iteration as an environment, the curricula generated by the meta-solver as action, and the negative exploitability as the reward for the meta-solver RL agent. 
In other words, we formulate the PSRO process as an independent MDP, similar to \cite{duan2016rl}.  
Following this idea, we reformulate the training of the meta-solver as an RL problem with a continuous action space and solve such an MDP with Deep Deterministic Policy Gradient (DDPG).

We conduct experiments to train the meta-solver with DDPG on 2D-RPS. Empirically we find that the trained meta-solver can achieve better performance compared with PSRO-Uniform. However, it cannot beat PSRO, unlike our meta-gradient based meta-solver. We believe that this might be because the dynamics of the PSRO environment is complicated which makes it rather challenging  for DDPG to learn a good policy (i.e., the meta-solver).
\begin{figure}[H]
    \centering
    \includegraphics[width=0.65\textwidth]{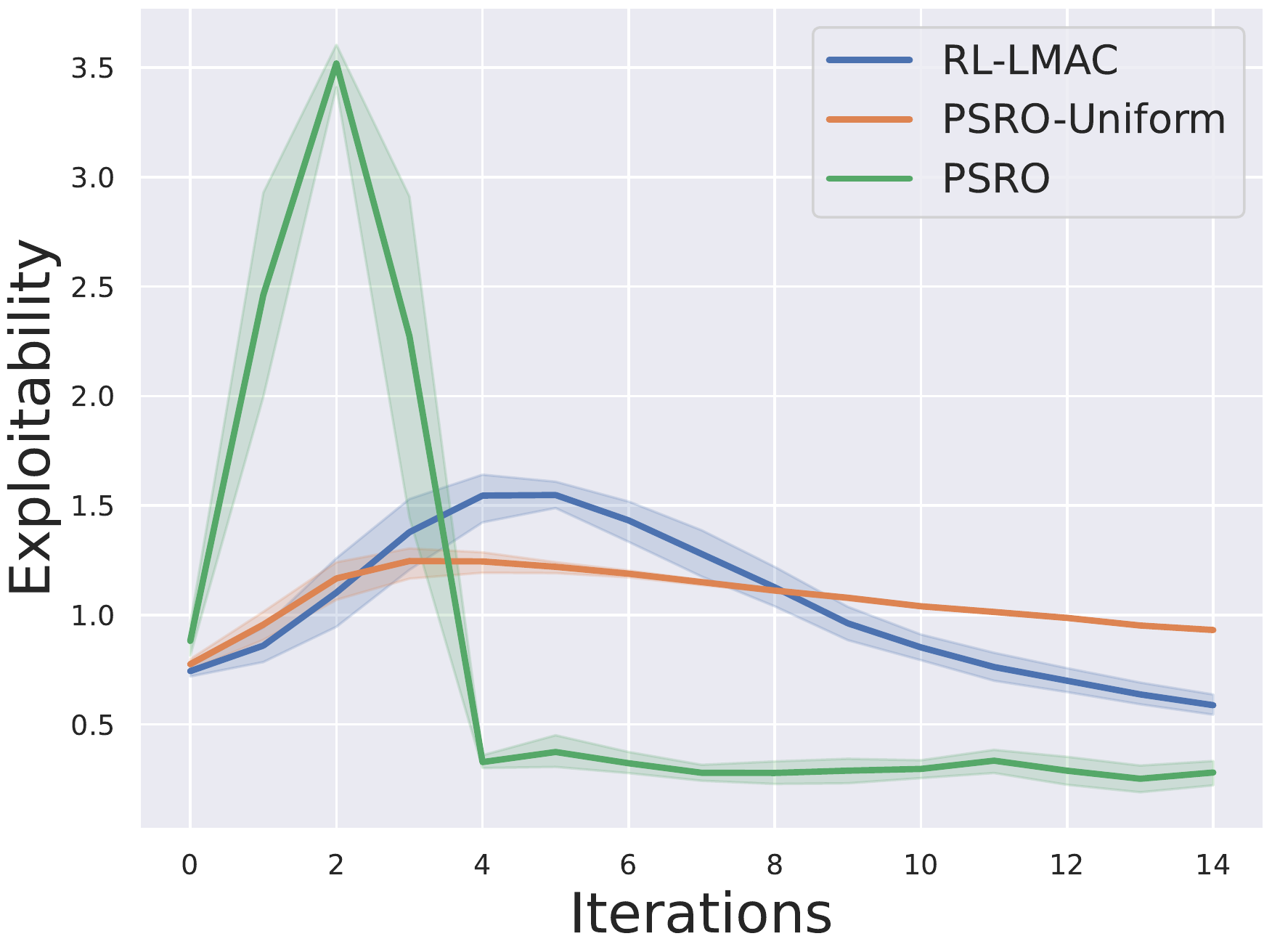}
    \caption{NAC trained with DDPG for environment on 2D-RPS.}
    \label{ap:spgen}
\end{figure}
\clearpage
\section{Additional Implementation Details}
We report any relevant additional implementation details in this section. \begin{tcolorbox}
The code can be found in: \url{https://github.com/waterhorse1/NAC}
\end{tcolorbox}\label{additional_implementaion}
\subsection{Environment Description}
\label{envir}
\textbf{Random Games of Skill}
\cite{czarnecki2020real} are normal-form games designed to consist of both a transitive and non-transitive element. The payoff function is shown as: $G_{i, j}:=\frac{1}{2}(W_{i, j}-W_{j, i})+S_{i}-S_{j} \text { with } W_{i, j}, S_{i} \stackrel{i . i . d}{\sim} \mathcal{N}(0, \sigma_{W}^{2} \text { or } \sigma_{S}^{2})$. The intuition behind random games of skill is to model the transitive strength of a strategy via $S$ and the non-transitive cycles by $W$. \cite{czarnecki2020real} shows that many real worlds game exhibits the geometry property of games of skill. 

In our experimental setting, we increase the presence of non-transitive cycles by substituting $\frac{1}{2}(W_{i, j}-W_{j, i})$ with $(W_{i, j}-W_{j, i})$. Note that random games of skill naturally provides us with a distribution $P(G)$ over games. We set the meta training distribution on 200*200 games of skill matrix and utilise gradient descent for best-response policy update and exploitability calculation in PSRO. This is a symmetric game so we only need to construct one policy pool for PSRO. Note that the best-response for exploitability calculation in GOS is gradient descent rather than direct maximisation over all pure actions, so it is actually an approximate exploitability. It might bring in negative exploitability but it is still a fair comparison because we use the same way to calculate the exploitability for all algorithms.

\textbf{Differentiable Lotto.}
Differentiable lotto is a game inspired by \cite{hart2008discrete} and is introduced in \cite{balduzzi2019open}. This game is defined over a fixed set of customers $c_i\in \mathbb{R}^{2}, i \in \{1,...,n\}$ where each customer represents a fixed point on a 2D plane. In this game, each agent determines $\left\{\left(p_{1}, \mathbf{v}_{1}\right), \ldots,\left(p_{k}, \mathbf{v}_{k}\right)\right\}$, where $v_i$ and $p_i$ respectively denote the position and the units of resources of server $i$. Given two agent $(\mathbf{p}, \mathbf{v})$ and $(\mathbf{q}, \mathbf{w})$, the customers c are softly assigned to servers based on the distance between customer and server. The payoff function is then given as: $
\phi((\mathbf{p}, \mathbf{v}),(\mathbf{q}, \mathbf{w})):=\sum_{i, j=1}^{c, k}\left(p_{j} v_{i j}-q_{j} w_{i j}\right)
\text{, where }\left(v_{i 1}, \ldots, w_{i k}\right):=\operatorname{softmax}(-\left\|\mathbf{c}_{i}-\mathbf{v}_{1}\right\|^{2}, \ldots,-\left\|\mathbf{c}_{i}-\mathbf{w}_{k}\right\|^{2})$. This game is a relatively more transitive game compared with Random Games of Skill and Non-transitive Mixture Model game. And since there exists infinite points over 2d plane, Differentiable Lotto is an open-ended game. 

In our experiments, we use 9 customers and 500 servers. We set meta-training distribution by randomizing the customers positions and the initial positions of servers according to $\mathcal{N}(0,1)$. Gradient descent is utilized for best-response policy update and exploitability calculation in PSRO. Note that this is a symmetric game so we only need to construct one policy pool for PSRO.

\textbf{Non-Transitive Mixture Model.}
Non-Transitive Mixture Model is also an open-ended game with both transitivity and non-transitivity. To achieve Nash policy, the player needs to not only climb up to the Gaussian distribution to maximize transitive payoff, but also explore each Gaussian distribution to remain un-exploitable. 

In our experiment, we set $n=7$, and randomize the center of Gaussian distribution and initial position of strategies for meta-training distribution. Gradient descent is utilized for best-response policy update and exploitability calculation in PSRO. Note that this is a symmetric game so we only need to construct one policy pool for PSRO.

\textbf{Iterated Matching Pennies (IMP).}
\begin{table}[H]
\vspace{-10pt}
    \caption{Matching pennies}
    \centering
    \begin{tabular}{c|c|c}
     \hline
     &  Head & Tail\\
     \hline
     Head&  (+a, -a) & (-a, +a)\\
     \hline
     Tail&  (-b, +b) & (+b, -b)\\
     \hline
    \end{tabular}
    \label{tab:imp}
    \vspace{-15pt}
\end{table}

We follow the works of \cite{foerster2017learning} and \cite{kim2020policy} in using  IMP \cite{gibbons1992game}, a zero-sum game in which the row player wants to have matching pennies whilst the column player wants to have clashing pennies. The original matching pennies game is shown in Table (\ref{tab:imp}) as $a=b=1$. We extend it to the iterated form where agents can condition their actions on past history. We follow \cite{foerster2017learning} to model it as a memory-1 two-agent MRP and agent's action at timestep $t$ will condition on the joint action at timestep $t-1$. As mentioned in Section (\ref{sec:meta_solver_upd}), our training framework, alongside most meta-learning frameworks, cannot tolerate a large amount of inner-loop gradient steps. Fortunately, IMP is fairly simple and does not need to take many policy gradient steps to reach an approximate best-response. We show Table (\ref{tab:imp}) as the stage-game of the iterated game played in IMP. We set $a,b \sim U(0.5, 2)$ as the meta-training distribution over the game. Policy gradient is utilized for best-response policy update and exploitability calculation in PSRO. The iteration length is 50.

In IMP, we follow the setting in \cite{foerster2017learning} where each agent's policy is fully specified by 5 probabilities. For agent
a in IMP, they are the probability of head
at game start $\pi^a(H|S_0)$, and the head probabilities in the
four memories: $\pi^a(H|HH)$, $\pi^a(H|HT)$, $\pi^a(H|TH)$ and $\pi^a(H|TT)$. Note that this is a non-symmetric game so we need to construct two policy pools for PSRO.

\textbf{Kuhn Poker}
was introduced by \cite{kuhnpoker} as a two-player, sequential-move, imperfect information poker game which has a total of 6 information states for each player, 12 overall. A round of Kuhn Poker is as follows: Both players start with 2 chips and both put in 1 chip in order to play. The deck is only 3 cards, and each player is dealt one card. At this point, both players have the choice of betting or passing - if both players take the same action then the player with the higher card wins, otherwise the player who made a bet wins. Kuhn Poker is a simplified version of poker which can be easily integrated with game theoretic analysis and is therefore well-aligned with the use of PSRO. Kuhn Poker has a large strategy space consisting $2^{12}$ pure strategies in total. 

In Kuhn Poker it is easy to find an exact best-response to a meta-strategy by traversing the game-tree and selecting the the action at each state with the highest expected value. The results of using this exact approach are shown in Appendix \ref{ap:kuhn-poker}. However, the central interest of our work is when using approximate best-responses, so we also suggest two different manners in which we can specify an approximate best-response when traversing the game-tree.

In Pseudo-code \ref{alg:tab-v1} we illustrate our first method, where the action with the highest expected value at each state will be played the majority of the time, but the policy may also take the action with the lower expected value with a lower probability. Notably, these action probability values are fixed at 0.75 and 0.25. We note that, whilst this setting performs well on the Kuhn Poker game, it is not able to generalise effectively to Leduc Poker.

In Pseudo-code \ref{alg:tab-v2} we illustrate our second method, where again the action with the highest expected value at each state will be played the majority of the time, but we introduce more randomness into the process. Strictly, we sample two perturbations $\boldsymbol{\eta}_1,\boldsymbol{\eta}_2 \sim \mathcal{N}(0, 1)$ and the action with the highest expected value will be played with probability $1 + \eta_1$ and the other action will be played with probability $\eta_2$. We believe this to be a fundamentally more applicable measure as, because in Kuhn Poker it is difficult to define a distribution over games (as there is only one Kuhn Poker game), we instead have this setting defined over the distribution of best-responses allowing us to maintain a distribution setting. We believe this distribution over best-responses is what allows this method to generalise well to Leduc Poker, as it is able to explore more dynamics of the game-type. 

\subsection{Implementation Details}
\label{implentation}
In this section we will list any specific implementation details that we used for each meta-solver training experiment.

\textbf{NAC With Gradient-Descent Best-Response Oracles}
\begin{itemize}
    \item In order to control for any instances of gradient explosion, we apply gradient clip normalisation on the meta-gradient with the clip parameter being reported in Appendix \ref{ap:hyper}.
    \item In order to speed up the training process, we distributed each game in a batch across multiple training nodes.
\end{itemize}

\textbf{NAC With Gradient-Descent Best-Response Oracles - Implicit}
\begin{itemize}
    \item In order to control for any instances of gradient explosion, we apply gradient clip normalisation on the meta-gradient with the clip parameter being reported in Appendix \ref{ap:hyper}.
    \item In order to meet the stationary point condition for implicit gradient, we take enough inner-loop gradient steps until the gradient norm is below the threshold.
    \item In order to speed up the training process, we distributed each game in a batch across multiple training nodes.
\end{itemize}

\textbf{NAC With Reinforcement Learning Best-Response Oracles}
\begin{itemize}
    \item In order to control for any instances of gradient explosion, we apply a special trick - layer-wise gradient normalisation on the meta-gradient with the clip parameter being reported in Appendix \ref{ap:hyper}.
    \item In order to speed up the training process, we distributed each game in a batch across multiple training nodes.
    \item We apply linear baseline method in the inner-loop rl based best-response for variance reduction. This is a commonly used strategy for reinforcement learning based meta learning\cite{finn2017model}.
\end{itemize}

\textbf{ES-NAC}
\begin{itemize}
    \item In order to speed up the training process, we distributed each perturbation of the meta-solver across multiple training nodes.
\end{itemize}
\subsection{Meta-testing}
\label{Meta-testing}
There exist some differences between the baseline algorithms and NAC when we conduct meta-testing. Since the baseline algorithms need no further training, in the testing phase, we evaluate the baseline algorithms on multiple tasks sampled from the task distribution so the confidence interval for baseline algorithms refers to the randomness brought by different tasks. However, since we need to conduct the training on multiple seeds for NAC (so there exist two random variables - task and random seed), we follow previous Meta-RL evaluation way \cite{rakelly2019efficient} to have each trained model tested on multiple tasks and calculate the mean exploitability over tasks. It can reduce the randomness brought by task distribution. So the confidence interval in the plot for NAC refers to the standard deviation brought by different random seeds.
\subsection{Computing Infrastructure}
\label{computing}
We used two internal compute servers both consisting of 4x Nvidia GeForce 1080-Ti cards, however each model is trained on at most 1 card. Additionally we made use of High Performance Computing Cluster for ES experiments.
\section{Hyperparameter Details}
\label{hyper}
 
We report our hyperparameter settings we use for experiments in this section.
\label{ap:hyper}

\subsection{Games of Skill - Alg. \ref{alg:auto-psro-gd-nonimp}}
\begin{table}[H]
\caption{Hyper-parameter settings for Random Games of Skill Training.}
\label{tb:hyp_gos}
\centering
\begin{sc}
\resizebox{1. \textwidth}{!}{
\begin{tabular}{lcl}
\toprule
\textbf{Settings }& \textbf{Value} & \textbf{Description}  \\  \hline \hline
Oracle method & Gradient Descent & subroutine of getting oracles  \\
Outer Learning rate & 0.01 & Learning rate for meta-solver updates  \\
Meta training Steps & 100 & Number of meta-solver update steps  \\
Meta Batch Size & 5 & Number of games trained on per iteration  \\
Model Type & GRU & Type of meta-solver  \\
Gradient Clip Value & 1.0 & Meta-gradient clip value \\
PSRO Iterations & 20 & Number of PSRO Iterations  \\
Window Size & 5 & Number of window size\\
Inner Learning rate & 25.0 & Learning rate for best-response updates  \\
Inner GD Steps & 5 & Number of best-response update steps  \\
Exploitability Learning rate & 10.0 & Learning rate for exploitability calculation  \\
Inner Exploitability Steps & 20 & Number of exploitability update steps  \\
\hline \hline
\bottomrule
\end{tabular}
}
\end{sc}
\end{table}
\subsection{Differentiable Blotto - Alg. \ref{alg:auto-psro-gd-nonimp}}
\begin{table}[H]
\caption{Hyper-parameter settings for Differentiable Lotto Training.}
\label{tb:hyp_lotto}
\centering
\begin{sc}
\resizebox{1. \textwidth}{!}{
\begin{tabular}{lcl}
\toprule
\textbf{Settings }& \textbf{Value} & \textbf{Description}  \\  \hline \hline
Oracle method & Gradient Descent & subroutine of getting oracles  \\
Outer Learning rate & 0.001 & Learning rate for meta-solver updates  \\
Meta training Steps & 100 & Number of meta-solver update steps  \\
Meta Batch Size & 5 & Number of games trained on per iteration  \\
Model Type & GRU & Type of meta-solver  \\
Gradient Clip Value & 1.0 & Meta-gradient clip value \\
PSRO Iterations & 20 & Number of PSRO Iterations  \\
Window Size & 5 & Number of window size\\
Inner Learning rate & 20.0 & Learning rate for best-response updates  \\
Inner GD Steps & 20 & Number of best-response update steps  \\
Exploitability Learning rate & 20.0 & Learning rate for exploitability calculation  \\
Inner Exploitability Steps & 30 & Number of exploitability update steps  \\
\hline \hline
\bottomrule
\end{tabular}
}
\end{sc}
\end{table}
\subsection{Non-transitive Mixture Model}
\subsubsection{Best response by Non-implicit Gradient Descent - Alg. \ref{alg:auto-psro-gd-nonimp}}
\begin{table}[H]
\caption{Hyper-parameter settings for non-implicit 2D-RPS Training.}
\label{tb:hyp_gos}
\centering
\begin{sc}
\resizebox{1. \textwidth}{!}{
\begin{tabular}{lcl}
\toprule
\textbf{Settings }& \textbf{Value} & \textbf{Description}  \\  \hline \hline
Oracle method & Gradient Descent & subroutine of getting oracles  \\
Outer Learning rate & 0.007 & Learning rate for meta-solver updates  \\
Meta training steps & 400 & Number of meta-solver update steps  \\
Meta Batch Size & 8 & Number of games trained on per iteration  \\
Model Type & Conv1D & Type of meta-solver  \\
LR Schedule Step & 100 & Outer LR Scheduler step iteration \\
LR Schedule Gamma & 0.3 & Outer LR Scheduler multiplicative value \\
Gradient Clip Value & 2.0 & Meta-gradient clip value \\
PSRO Iterations & 15 & Number of PSRO Iterations  \\
Window Size & 9 & Number of window size\\
Inner Learning rate & 2.0 & Learning rate for best-response updates  \\
Inner GD Steps & 5 & Number of best-response update steps  \\
Exploitability Learning rate & 2.0 & Learning rate for exploitability calculation  \\
Inner Exploitability Steps & 20 & Number of exploitability update steps  \\
\hline \hline
\bottomrule
\end{tabular}
}
\end{sc}
\end{table}
\subsubsection{Best response by Implicit Gradient Descent - Alg. \ref{alg:auto-psro-gd-imp}}
\begin{table}[H]
\caption{Hyper-parameter settings for implicit 2D-RPS Training.}
\label{tb:hyp_gos}
\centering
\begin{sc}
\resizebox{1. \textwidth}{!}{
\begin{tabular}{lcl}
\toprule
\textbf{Settings }& \textbf{Value} & \textbf{Description}  \\  \hline \hline
Oracle method & Gradient Descent & subroutine of getting oracles  \\
Outer Learning rate & 0.005 & Learning rate for meta-solver updates  \\
Meta training steps & 600 & Number of meta-solver update steps  \\
Meta Batch Size & 10 & Number of games trained on per iteration  \\
Model Type & Conv1D & Type of meta-solver  \\
Gradient Clip Values & 0.002 & Value above which meta-gradient is clipped \\
PSRO Iterations & 10 & Number of PSRO Iterations  \\
Window Size & 10 & Number of window size\\
Inner Learning rate & 0.75 & Learning rate for best-response updates  \\
Inner GD Steps & 100 & Number of best-response update steps  \\
Exploitability Learning rate & 0.75 & Learning rate for exploitability calculation  \\
Inner Exploitability Steps & 200 & Number of exploitability update steps  \\
Inner-loop Gradient Norm Break Value & 0.001 & Value at which inner-loop gradient update is stopped. \\
\hline \hline
\bottomrule
\end{tabular}
}
\end{sc}
\end{table}
\subsection{Iterated Matching Pennies - Alg. \ref{alg:auto-psro-rl}}
\begin{table}[H]
\caption{Hyper-parameter settings for Iterated Matching Pennies Training.}
\label{tb:hyp_gos}
\centering
\begin{sc}
\resizebox{1. \textwidth}{!}{
\begin{tabular}{lcl}
\toprule
\textbf{Settings }& \textbf{Value} & \textbf{Description}  \\  \hline \hline
Oracle method & Reinforce & subroutine of getting oracles  \\
Outer Learning rate & 0.004 & Learning rate for meta-solver updates  \\
Meta Training Steps & 50 & Number of meta-solver update steps  \\
Meta Batch Size & 8 & Number of games trained on per iteration  \\
Model Type & GRU & Type of meta-solver  \\
Layer-wise gradient normalisation threshold & 0.002 & Value above which layer-wise meta-gradient is clipped \\
PSRO Iterations & 9 & Number of PSRO Iterations  \\
Window Size & 3 & Number of window size\\
Inner Learning rate & 10.0 & Learning rate for best-response updates  \\
Inner GD Steps & 10 & Number of best-response update steps  \\
Exploitability Learning rate & 10.0 & Learning rate for exploitability calculation  \\
Exploitability Steps & 20 & Number of exploitability update steps  \\
Trajectories sampled each update & 32 & Number of trajectories sampled each reinforce update \\
\hline \hline
\bottomrule
\end{tabular}
}
\end{sc}
\end{table}
\subsection{Kuhn Poker}\label{app:kp}
\subsubsection{Best response by Approximate Tabular V1 - Alg. \ref{alg:tab-v1}}
\begin{table}[H]
\caption{Hyper-parameter settings for Kuhn Poker Tabular V1 Training.}
\label{tb:hyp_gos}
\centering
\begin{sc}
\resizebox{1. \textwidth}{!}{
\begin{tabular}{lcl}
\toprule
\textbf{Settings }& \textbf{Value} & \textbf{Description}  \\  \hline \hline
Oracle method & Approximate Tabular V1 & subroutine of getting oracles  \\
Outer Learning rate & 0.1 & Learning rate for meta-solver updates  \\
Meta Training Steps & 100 & Number of meta-solver update steps  \\
Meta Batch Size & 5 & Number of games trained on per iteration  \\
odel Type & Conv1D & Type of meta-solver  \\
LR Schedule Step & 50 & Outer LR Scheduler step iteration \\
LR Schedule Gamma & 0.5 & Outer LR Scheduler multiplicative value \\
PSRO Iterations & 15 & Number of PSRO Iterations  \\
ES Perturbations & 30 & Number of model perturbations via ES \\
\hline \hline
\bottomrule
\end{tabular}
}
\end{sc}
\end{table}
\subsubsection{Best response by Approximate Tabular V2 - Alg. \ref{alg:tab-v2}}
\begin{table}[H]
\caption{Hyper-parameter settings for Kuhn Poker Tabular V2 Training.}
\label{tb:hyp_gos}
\centering
\begin{sc}
\resizebox{1. \textwidth}{!}{
\begin{tabular}{lcl}
\toprule
\textbf{Settings }& \textbf{Value} & \textbf{Description}  \\  \hline \hline
Oracle method & Approximate Tabular V2 & subroutine of getting oracles  \\
Outer Learning rate & 0.1 & Learning rate for meta-solver updates  \\
Meta Training Steps & 100 & Number of meta-solver update steps  \\
Meta Batch Size & 5 & Number of games trained on per iteration  \\
Model Type & Conv1D & Type of meta-solver  \\
LR Schedule Step & 20 & Outer LR Scheduler step iteration \\
LR Schedule Gamma & 0.5 & Outer LR Scheduler multiplicative value \\
PSRO Iterations & 15 & Number of PSRO Iterations  \\
ES Perturbations & 30 & Number of model perturbations via ES \\
\hline \hline
\bottomrule
\end{tabular}
}
\end{sc}
\end{table}
\subsubsection{Best response by PPO}
\begin{table}[H]
\caption{Hyper-parameter settings for Kuhn Poker PPO Training.}
\label{tb:hyp_gos}
\centering
\begin{sc}
\resizebox{1. \textwidth}{!}{
\begin{tabular}{lcl}
\toprule
\textbf{Settings }& \textbf{Value} & \textbf{Description}  \\  \hline \hline
Oracle method & PPO & subroutine of getting oracles  \\
Outer Learning rate & 0.2 & Learning rate for meta-solver updates  \\
Meta Training Steps & 100 & Number of meta-solver update steps  \\
Meta Batch Size & 3 & Number of games trained on per iteration  \\
Model Type & Conv1D & Type of meta-solver  \\
LR Schedule Step & 20 & Outer LR Scheduler step iteration \\
LR Schedule Gamma & 0.5 & Outer LR Scheduler multiplicative value \\
ES Perturbations & 30 & Number of model perturbations via ES \\
PSRO Iterations & 10 & Number of PSRO Iterations  \\
PPO Clip Ratio & 0.8 & Clip Ratio of PPO Trainer \\
Pi LR & 0.003 & LR for policy optimiser \\
VF LR & 0.001 & LR for value function optimiser \\ 
Pi Train Iters & 100 & Number of policy optimiser training iters. \\
VF Train Iters & 100 & Number of VF optimiser training iters. \\
Target KL & 0.5 & Early Stopping Criteria \\
\hline \hline
\bottomrule
\end{tabular}
}
\end{sc}
\end{table}
\section{Author Contributions}
We summarise the main contributions from each of the authors as follows:

\textbf{Xidong Feng}: Idea proposing, algorithm design, code implementation and experiments running (on 2D-RPS, 2D-RPS-Implicit and IMP), and paper writing.

\textbf{Oliver Slumbers}: Algorithm design, code implementation and experiments running (on Gos, Blotto, Kuhn-Poker), and paper writing.

\textbf{Ziyu Wan}: Code implementation and experiments running for RL based NAC in Appendix \ref{ap:train_rl}.

\textbf{Bo Liu}: Experiments running for Kuhn-Poker.

\textbf{Stephen McAleer}: Project discussion and paper writing.

\textbf{Ying Wen}: Project discussion.

\textbf{Jun Wang}: Project discussion and overall project supervision.

\textbf{Yaodong Yang}: Project lead, idea proposing, experiment supervision, and paper writing.
\end{document}

%% file: math_commands.tex
\usepackage{amsmath,amsfonts,bm}

\def\eqref#1{equation~\ref{#1}}

\def\1{\bm{1}}

\def\vtheta{{\bm{\theta}}}

\def\vpi{{\bm{\pi}}}
\def\vPhi{{\bm{\Phi}}}

\def\mA{{\bm{A}}}

\def\mG{{\bm{G}}}

\def\mS{{\bm{S}}}

\DeclareMathAlphabet{\mathsfit}{\encodingdefault}{\sfdefault}{m}{sl}
\SetMathAlphabet{\mathsfit}{bold}{\encodingdefault}{\sfdefault}{bx}{n}
\newcommand{\tens}[1]{\bm{\mathsfit{#1}}}

\def\tM{{\tens{M}}}

\DeclareMathOperator*{\argmax}{arg\,max}
\DeclareMathOperator*{\argmin}{arg\,min}

%% file: arxiv.bbl
\begin{thebibliography}{10}

\bibitem{al2017continuous}
Maruan Al-Shedivat, Trapit Bansal, Yuri Burda, Ilya Sutskever, Igor Mordatch,
  and Pieter Abbeel.
\newblock Continuous adaptation via meta-learning in nonstationary and
  competitive environments.
\newblock {\em arXiv preprint arXiv:1710.03641}, 2017.

\bibitem{baker2019emergent}
Bowen Baker, Ingmar Kanitscheider, Todor Markov, Yi~Wu, Glenn Powell, Bob
  McGrew, and Igor Mordatch.
\newblock Emergent tool use from multi-agent autocurricula.
\newblock {\em arXiv preprint arXiv:1909.07528}, 2019.

\bibitem{balduzzi2019open}
D~Balduzzi, M~Garnelo, Y~Bachrach, W~Czarnecki, J~P{\'e}rolat, M~Jaderberg, and
  T~Graepel.
\newblock Open-ended learning in symmetric zero-sum games.
\newblock In {\em ICML}, volume~97, pages 434--443. PMLR, 2019.

\bibitem{bechtle2021meta}
Sarah Bechtle, Artem Molchanov, Yevgen Chebotar, Edward Grefenstette, Ludovic
  Righetti, Gaurav Sukhatme, and Franziska Meier.
\newblock Meta learning via learned loss.
\newblock In {\em 2020 25th International Conference on Pattern Recognition
  (ICPR)}, pages 4161--4168. IEEE, 2021.

\bibitem{brown1951iterative}
George~W Brown.
\newblock Iterative solution of games by fictitious play.
\newblock {\em Activity analysis of production and allocation}, 13(1):374--376,
  1951.

\bibitem{choromanski2018structured}
Krzysztof Choromanski, Mark Rowland, Vikas Sindhwani, Richard~E. Turner, and
  Adrian Weller.
\newblock Structured evolution with compact architectures for scalable policy
  optimization, 2018.

\bibitem{chung2014empirical}
Junyoung Chung, Caglar Gulcehre, KyungHyun Cho, and Yoshua Bengio.
\newblock Empirical evaluation of gated recurrent neural networks on sequence
  modeling.
\newblock {\em arXiv preprint arXiv:1412.3555}, 2014.

\bibitem{czarnecki2020real}
Wojciech~Marian Czarnecki, Gauthier Gidel, Brendan Tracey, Karl Tuyls, Shayegan
  Omidshafiei, David Balduzzi, and Max Jaderberg.
\newblock Real world games look like spinning tops.
\newblock {\em arXiv preprint arXiv:2004.09468}, 2020.

\bibitem{davis2014using}
Trevor Davis, Neil Burch, and Michael Bowling.
\newblock Using response functions to measure strategy strength.
\newblock In {\em Proceedings of the AAAI Conference on Artificial
  Intelligence}, volume~28, 2014.

\bibitem{deng2021complexity}
Xiaotie Deng, Yuhao Li, David~Henry Mguni, Jun Wang, and Yaodong Yang.
\newblock On the complexity of computing markov perfect equilibrium in
  general-sum stochastic games.
\newblock {\em arXiv preprint arXiv:2109.01795}, 2021.

\bibitem{dinh2021online}
Le~Cong Dinh, Yaodong Yang, Zheng Tian, Nicolas~Perez Nieves, Oliver Slumbers,
  David~Henry Mguni, and Jun Wang.
\newblock Online double oracle.
\newblock {\em arXiv preprint arXiv:2103.07780}, 2021.

\bibitem{duan2016rl}
Yan Duan, John Schulman, Xi~Chen, Peter~L Bartlett, Ilya Sutskever, and Pieter
  Abbeel.
\newblock Rl: Fast reinforcement learning via slow reinforcement learning.
\newblock {\em arXiv preprint arXiv:1611.02779}, 2016.

\bibitem{finn2017model}
Chelsea Finn, Pieter Abbeel, and Sergey Levine.
\newblock Model-agnostic meta-learning for fast adaptation of deep networks.
\newblock In {\em International Conference on Machine Learning}, pages
  1126--1135. PMLR, 2017.

\bibitem{foerster2018dice}
Jakob Foerster, Gregory Farquhar, Maruan Al-Shedivat, Tim Rockt{\"a}schel, Eric
  Xing, and Shimon Whiteson.
\newblock Dice: The infinitely differentiable monte carlo estimator.
\newblock In {\em International Conference on Machine Learning}, pages
  1529--1538. PMLR, 2018.

\bibitem{foerster2017learning}
Jakob~N Foerster, Richard~Y Chen, Maruan Al-Shedivat, Shimon Whiteson, Pieter
  Abbeel, and Igor Mordatch.
\newblock Learning with opponent-learning awareness.
\newblock {\em arXiv preprint arXiv:1709.04326}, 2017.

\bibitem{gibbons1992game}
Robert~S Gibbons.
\newblock {\em Game theory for applied economists}.
\newblock Princeton University Press, 1992.

\bibitem{hart2008discrete}
Sergiu Hart.
\newblock Discrete colonel blotto and general lotto games.
\newblock {\em International Journal of Game Theory}, 36(3):441--460, 2008.

\bibitem{hessel2018rainbow}
Matteo Hessel, Joseph Modayil, Hado Van~Hasselt, Tom Schaul, Georg Ostrovski,
  Will Dabney, Dan Horgan, Bilal Piot, Mohammad Azar, and David Silver.
\newblock Rainbow: Combining improvements in deep reinforcement learning.
\newblock In {\em Proceedings of the AAAI Conference on Artificial
  Intelligence}, volume~32, 2018.

\bibitem{hochreiter1997long}
Sepp Hochreiter and J{\"u}rgen Schmidhuber.
\newblock Long short-term memory.
\newblock {\em Neural computation}, 9(8):1735--1780, 1997.

\bibitem{houthooft2018evolved}
Rein Houthooft, Richard~Y Chen, Phillip Isola, Bradly~C Stadie, Filip Wolski,
  Jonathan Ho, and Pieter Abbeel.
\newblock Evolved policy gradients.
\newblock {\em arXiv preprint arXiv:1802.04821}, 2018.

\bibitem{kim2020policy}
Dong-Ki Kim, Miao Liu, Matthew Riemer, Chuangchuang Sun, Marwa Abdulhai, Golnaz
  Habibi, Sebastian Lopez-Cot, Gerald Tesauro, and Jonathan~P How.
\newblock A policy gradient algorithm for learning to learn in multiagent
  reinforcement learning.
\newblock {\em arXiv preprint arXiv:2011.00382}, 2020.

\bibitem{kirsch2019improving}
Louis Kirsch, Sjoerd van Steenkiste, and Juergen Schmidhuber.
\newblock Improving generalization in meta reinforcement learning using learned
  objectives.
\newblock In {\em International Conference on Learning Representations}, 2019.

\bibitem{kuhnpoker}
Harold~W Kuhn.
\newblock 9. a simplified two-person poker.
\newblock In {\em Contributions to the Theory of Games (AM-24), Volume I},
  pages 97--104. Princeton University Press, 2016.

\bibitem{lanctot2019openspiel}
Marc Lanctot, Edward Lockhart, Jean-Baptiste Lespiau, Vinicius Zambaldi,
  Satyaki Upadhyay, Julien P{\'e}rolat, Sriram Srinivasan, Finbarr Timbers,
  Karl Tuyls, Shayegan Omidshafiei, et~al.
\newblock Openspiel: A framework for reinforcement learning in games.
\newblock {\em arXiv preprint arXiv:1908.09453}, 2019.

\bibitem{lanctot2017unified}
Marc Lanctot, Vinicius Zambaldi, Audrunas Gruslys, Angeliki Lazaridou, Karl
  Tuyls, Julien Perolat, David Silver, and Thore Graepel.
\newblock A unified game-theoretic approach to multiagent reinforcement
  learning, 2017.

\bibitem{leibo2019autocurricula}
Joel~Z Leibo, Edward Hughes, Marc Lanctot, and Thore Graepel.
\newblock Autocurricula and the emergence of innovation from social
  interaction: A manifesto for multi-agent intelligence research.
\newblock {\em arXiv e-prints}, pages arXiv--1903, 2019.

\bibitem{liu2019taming}
Hao Liu, Richard Socher, and Caiming Xiong.
\newblock Taming maml: Efficient unbiased meta-reinforcement learning.
\newblock In {\em International Conference on Machine Learning}, pages
  4061--4071. PMLR, 2019.

\bibitem{liu2021unifying}
Xiangyu Liu, Hangtian Jia, Ying Wen, Yaodong Yang, Yujing Hu, Yingfeng Chen,
  Changjie Fan, and Zhipeng Hu.
\newblock Unifying behavioral and response diversity for open-ended learning in
  zero-sum games.
\newblock {\em arXiv preprint arXiv:2106.04958}, 2021.

\bibitem{long2015fully}
Jonathan Long, Evan Shelhamer, and Trevor Darrell.
\newblock Fully convolutional networks for semantic segmentation, 2015.

\bibitem{lorraine2020optimizing}
Jonathan Lorraine, Paul Vicol, and David Duvenaud.
\newblock Optimizing millions of hyperparameters by implicit differentiation.
\newblock In {\em International Conference on Artificial Intelligence and
  Statistics}, pages 1540--1552. PMLR, 2020.

\bibitem{mcaleer2021xdo}
Stephen McAleer, John Lanier, Pierre Baldi, and Roy Fox.
\newblock {XDO}: A double oracle algorithm for extensive-form games.
\newblock {\em Reinforcement Learning in Games Workshop, AAAI}, 2021.

\bibitem{mcaleer2020pipeline}
Stephen McAleer, John Lanier, Roy Fox, and Pierre Baldi.
\newblock Pipeline {PSRO}: A scalable approach for finding approximate nash
  equilibria in large games.
\newblock In {\em Advances in Neural Information Processing Systems (NeurIPS)},
  2020.

\bibitem{mcmahan2003planning}
H~Brendan McMahan, Geoffrey~J Gordon, and Avrim Blum.
\newblock Planning in the presence of cost functions controlled by an
  adversary.
\newblock In {\em Proceedings of the 20th International Conference on Machine
  Learning (ICML-03)}, pages 536--543, 2003.

\bibitem{morgenstern1953theory}
Oskar Morgenstern and John Von~Neumann.
\newblock {\em Theory of games and economic behavior}.
\newblock Princeton university press, 1953.

\bibitem{muller2019generalized}
Paul Muller, Shayegan Omidshafiei, Mark Rowland, Karl Tuyls, Julien Perolat,
  Siqi Liu, Daniel Hennes, Luke Marris, Marc Lanctot, Edward Hughes, et~al.
\newblock A generalized training approach for multiagent learning.
\newblock In {\em International Conference on Learning Representations}, 2019.

\bibitem{nash1950equilibrium}
John~F Nash et~al.
\newblock Equilibrium points in n-person games.
\newblock {\em Proceedings of the national academy of sciences}, 36(1):48--49,
  1950.

\bibitem{nieves2021modelling}
Nicolas~Perez Nieves, Yaodong Yang, Oliver Slumbers, David~Henry Mguni, and Jun
  Wang.
\newblock Modelling behavioural diversity for learning in open-ended games.
\newblock {\em arXiv preprint arXiv:2103.07927}, 2021.

\bibitem{oh2020discovering}
Junhyuk Oh, Matteo Hessel, Wojciech~M Czarnecki, Zhongwen Xu, Hado van Hasselt,
  Satinder Singh, and David Silver.
\newblock Discovering reinforcement learning algorithms.
\newblock {\em arXiv preprint arXiv:2007.08794}, 2020.

\bibitem{qi2017pointnet}
Charles~R Qi, Hao Su, Kaichun Mo, and Leonidas~J Guibas.
\newblock Pointnet: Deep learning on point sets for 3d classification and
  segmentation.
\newblock In {\em Proceedings of the IEEE conference on computer vision and
  pattern recognition}, pages 652--660, 2017.

\bibitem{rajeswaran2019metalearning}
Aravind Rajeswaran, Chelsea Finn, Sham Kakade, and Sergey Levine.
\newblock Meta-learning with implicit gradients, 2019.

\bibitem{rakelly2019efficient}
Kate Rakelly, Aurick Zhou, Chelsea Finn, Sergey Levine, and Deirdre Quillen.
\newblock Efficient off-policy meta-reinforcement learning via probabilistic
  context variables.
\newblock In {\em International conference on machine learning}, pages
  5331--5340. PMLR, 2019.

\bibitem{rothfuss2018promp}
Jonas Rothfuss, Dennis Lee, Ignasi Clavera, Tamim Asfour, and Pieter Abbeel.
\newblock Promp: Proximal meta-policy search.
\newblock {\em arXiv preprint arXiv:1810.06784}, 2018.

\bibitem{salimans2017evolution}
Tim Salimans, Jonathan Ho, Xi~Chen, Szymon Sidor, and Ilya Sutskever.
\newblock Evolution strategies as a scalable alternative to reinforcement
  learning, 2017.

\bibitem{sanjaya2021measuring}
Ricky Sanjaya, Jun Wang, and Yaodong Yang.
\newblock Measuring the non-transitivity in chess, 2021.

\bibitem{schulman2017proximal}
John Schulman, Filip Wolski, Prafulla Dhariwal, Alec Radford, and Oleg Klimov.
\newblock Proximal policy optimization algorithms.
\newblock {\em arXiv preprint arXiv:1707.06347}, 2017.

\bibitem{shao2021credit}
Jianzhun Shao, Hongchang Zhang, Yuhang Jiang, Shuncheng He, and Xiangyang Ji.
\newblock Credit assignment with meta-policy gradient for multi-agent
  reinforcement learning.
\newblock {\em arXiv preprint arXiv:2102.12957}, 2021.

\bibitem{singh2020parrot}
Avi Singh, Huihan Liu, Gaoyue Zhou, Albert Yu, Nicholas Rhinehart, and Sergey
  Levine.
\newblock Parrot: Data-driven behavioral priors for reinforcement learning.
\newblock {\em arXiv preprint arXiv:2011.10024}, 2020.

\bibitem{song2019maml}
Xingyou Song, Wenbo Gao, Yuxiang Yang, Krzysztof Choromanski, Aldo Pacchiano,
  and Yunhao Tang.
\newblock Es-maml: Simple hessian-free meta learning.
\newblock {\em arXiv preprint arXiv:1910.01215}, 2019.

\bibitem{sunehag2017value}
Peter Sunehag, Guy Lever, Audrunas Gruslys, Wojciech~Marian Czarnecki, Vinicius
  Zambaldi, Max Jaderberg, Marc Lanctot, Nicolas Sonnerat, Joel~Z Leibo, Karl
  Tuyls, et~al.
\newblock Value-decomposition networks for cooperative multi-agent learning.
\newblock {\em arXiv preprint arXiv:1706.05296}, 2017.

\bibitem{team2021open}
Ended~Learning Team, Adam Stooke, Anuj Mahajan, Catarina Barros, Charlie Deck,
  Jakob Bauer, Jakub Sygnowski, Maja Trebacz, Max Jaderberg, Michael Mathieu,
  et~al.
\newblock Open-ended learning leads to generally capable agents.
\newblock {\em arXiv preprint arXiv:2107.12808}, 2021.

\bibitem{van2020deterministic}
Jan van~den Brand.
\newblock A deterministic linear program solver in current matrix
  multiplication time.
\newblock In {\em Proceedings of the Fourteenth Annual ACM-SIAM Symposium on
  Discrete Algorithms}, pages 259--278. SIAM, 2020.

\bibitem{veeriah2019discovery}
Vivek Veeriah, Matteo Hessel, Zhongwen Xu, Janarthanan Rajendran, Richard~L
  Lewis, Junhyuk Oh, Hado van Hasselt, David Silver, and Satinder Singh.
\newblock Discovery of useful questions as auxiliary tasks.
\newblock In {\em NeurIPS}, 2019.

\bibitem{vinyals2019grandmaster}
Oriol Vinyals, Igor Babuschkin, Wojciech~M Czarnecki, Micha{\"e}l Mathieu,
  Andrew Dudzik, Junyoung Chung, David~H Choi, Richard Powell, Timo Ewalds,
  Petko Georgiev, et~al.
\newblock Grandmaster level in starcraft ii using multi-agent reinforcement
  learning.
\newblock {\em Nature}, 575(7782):350--354, 2019.

\bibitem{wang2016learning}
Jane~X Wang, Zeb Kurth-Nelson, Dhruva Tirumala, Hubert Soyer, Joel~Z Leibo,
  Remi Munos, Charles Blundell, Dharshan Kumaran, and Matt Botvinick.
\newblock Learning to reinforcement learn.
\newblock {\em arXiv preprint arXiv:1611.05763}, 2016.

\bibitem{williams1992simple}
Ronald~J Williams.
\newblock Simple statistical gradient-following algorithms for connectionist
  reinforcement learning.
\newblock {\em Machine learning}, 8(3-4):229--256, 1992.

\bibitem{xu2020meta}
Zhongwen Xu, Hado van Hasselt, Matteo Hessel, Junhyuk Oh, Satinder Singh, and
  David Silver.
\newblock Meta-gradient reinforcement learning with an objective discovered
  online.
\newblock {\em arXiv preprint arXiv:2007.08433}, 2020.

\bibitem{xu2018meta}
Zhongwen Xu, Hado van Hasselt, and David Silver.
\newblock Meta-gradient reinforcement learning.
\newblock {\em arXiv preprint arXiv:1805.09801}, 2018.

\bibitem{yang2021diverse}
Yaodong Yang, Jun Luo, Ying Wen, Oliver Slumbers, Daniel Graves, Haitham
  Bou~Ammar, Jun Wang, and Matthew~E Taylor.
\newblock Diverse auto-curriculum is critical for successful real-world
  multiagent learning systems.
\newblock In {\em Proceedings of the 20th International Conference on
  Autonomous Agents and MultiAgent Systems}, pages 51--56, 2021.

\bibitem{yang2020alphaalpha}
Yaodong Yang, Rasul Tutunov, Phu Sakulwongtana, and Haitham~Bou Ammar.
\newblock $\alpha$$\alpha$-rank: Practically scaling $\alpha$-rank through
  stochastic optimisation.
\newblock In {\em Proceedings of the 19th International Conference on
  Autonomous Agents and MultiAgent Systems}, pages 1575--1583, 2020.

\bibitem{yang2020overview}
Yaodong Yang and Jun Wang.
\newblock An overview of multi-agent reinforcement learning from game
  theoretical perspective.
\newblock {\em arXiv preprint arXiv:2011.00583}, 2020.

\bibitem{yang2020multi}
Yaodong Yang, Ying Wen, Jun Wang, Liheng Chen, Kun Shao, David Mguni, and
  Weinan Zhang.
\newblock Multi-agent determinantal q-learning.
\newblock In {\em International Conference on Machine Learning}, pages
  10757--10766. PMLR, 2020.

\bibitem{zahavy2020self}
Tom Zahavy, Zhongwen Xu, Vivek Veeriah, Matteo Hessel, Junhyuk Oh, Hado van
  Hasselt, David Silver, and Satinder Singh.
\newblock A self-tuning actor-critic algorithm.
\newblock {\em arXiv preprint arXiv:2002.12928}, 2020.

\bibitem{zaheer2017deep}
Manzil Zaheer, Satwik Kottur, Siamak Ravanbakhsh, Barnabas Poczos, Ruslan
  Salakhutdinov, and Alexander Smola.
\newblock Deep sets.
\newblock {\em arXiv preprint arXiv:1703.06114}, 2017.

\bibitem{zheng2020can}
Zeyu Zheng, Junhyuk Oh, Matteo Hessel, Zhongwen Xu, Manuel Kroiss, Hado
  Van~Hasselt, David Silver, and Satinder Singh.
\newblock What can learned intrinsic rewards capture?
\newblock In {\em International Conference on Machine Learning}, pages
  11436--11446. PMLR, 2020.

\bibitem{zheng2018learning}
Zeyu Zheng, Junhyuk Oh, and Satinder Singh.
\newblock On learning intrinsic rewards for policy gradient methods.
\newblock {\em Advances in Neural Information Processing Systems},
  31:4644--4654, 2018.

\bibitem{zhou2020online}
Wei Zhou, Yiying Li, Yongxin Yang, Huaimin Wang, and Timothy Hospedales.
\newblock Online meta-critic learning for off-policy actor-critic methods.
\newblock {\em Advances in Neural Information Processing Systems}, 33, 2020.

\end{thebibliography}
